\documentclass[journal,twoside]{IEEEtran}

\usepackage[hidelinks]{hyperref}
\usepackage{graphicx} 
\graphicspath{{./figures/}}
\usepackage{amsmath} 
\usepackage{amssymb}  
\usepackage{bbm}
\usepackage{bm}
\usepackage{verbatim} 
\usepackage{algpseudocode}
\usepackage{tabulary}
\usepackage{algorithm} 
\usepackage[utf8]{inputenc} 
\usepackage{wrapfig}
\usepackage{booktabs}
\usepackage{multirow}
\usepackage{caption}
\usepackage{color, colortbl}
\definecolor{Gray}{gray}{0.9}
\usepackage{xspace}
\usepackage{tcolorbox}

\newcounter{messagecounter}
\newcommand{\messagecounter}{\arabic{messagecounter}}
\newenvironment{messagebox}[1]{\begin{tcolorbox}[colback=black!15!white,colframe=white]\refstepcounter{messagecounter}{\textbf{Message \messagecounter:}}\label{#1}}{\end{tcolorbox}}

\makeatletter
\def\IfClass#1#2#3{\@ifundefined{opt@#1.cls}{#3}{#2}}
\makeatother

\IfClass{IEEEtran}{
\usepackage{cite}
}{}

\IfClass{IEEEtran}{

\newcommand{\citet}[1]{\cite{#1}}
}{}

\newcommand{\para}[1]{{\vspace{5mm}\noindent\textbf{#1}\hspace{0.3cm}}}

\usepackage{color}
\DeclareMathOperator*{\argmax}{\arg\!\max}
\DeclareMathOperator*{\argmin}{\arg\!\min}

\begin{document}

\title{A survey on policy search algorithms\\for learning robot controllers in a handful of trials}

\IfClass{IEEEtran}{
	\author{Konstantinos Chatzilygeroudis$^\dag$,
		Vassilis Vassiliades$^{\dag *}$,
		Freek Stulp$^\ddag$,
		Sylvain Calinon$^\diamond$
		and Jean-Baptiste Mouret$^\dag$
		\thanks{$^\dag$Inria, CNRS, Universit\'e de Lorraine, LORIA, F-54000 Nancy, France}
		\thanks{$^*$Research Centre on Interactive Media, Smart Systems and Emerging Technologies, Dimarcheio Lefkosias, Plateia Eleftherias, 1500, Nicosia, Cyprus}
		\thanks{$^\ddag$German Aerospace Center (DLR), Institute of Robotics and Mechatronics, Wessling, Germany}
		\thanks{$^\diamond$Idiap Research Institute, Rue Marconi 19, 1920 Martigny, Switzerland}
		\thanks{© 2019 IEEE.  Personal use of this material is permitted.  Permission from IEEE must be obtained for all other uses, in any current or future media, including reprinting/republishing this material for advertising or promotional purposes, creating new collective works, for resale or redistribution to servers or lists, or reuse of any copyrighted component of this work in other works.}}
	\markboth{IEEE Transactions on Robotics. Preprint version.}{Chatzilygeroudis \MakeLowercase{\textit{et al.}}: A survey on policy search algorithms for learning robot controllers in a handful of trials}
}{
	\titlesize{19}
	\author{Konstantinos Chatzilygeroudis$^\dag$,
		Vassilis Vassiliades$^\dag$,\\
		Freek Stulp$^\ddag$,
		Sylvain Calinon$^\diamond$
		and Jean-Baptiste Mouret$^\dag$}
	\authorshort{Chatzilygeroudis, Vassiliades, Stulp, Calinon and Mouret}
	\affiliations{$^\dag$Inria, CNRS, Universit\'e de Lorraine, LORIA, F-54000 Nancy, France\\
		$^\ddag$German Aerospace Center (DLR), Institute of Robotics and Mechatronics, Wessling, Germany\\
		$^\diamond$Idiap Research Institute, Martigny, Switzerland}
}

\maketitle


\begin{abstract}
	Most policy search algorithms require thousands of training episodes to find an effective policy, which is often infeasible with a physical robot. This survey article focuses on the extreme other end of the spectrum: how can a robot adapt with only a handful of trials (a dozen) and a few minutes? By analogy with the word ``big-data'', we refer to this challenge as ``micro-data reinforcement learning''. We show that a first strategy is to leverage prior knowledge on the policy structure (e.g., dynamic movement primitives), on the policy parameters (e.g., demonstrations), or on the dynamics (e.g., simulators). A second strategy is to create data-driven surrogate models of the expected reward (e.g., Bayesian optimization) or the dynamical model (e.g., model-based policy search), so that the policy optimizer queries the model instead of the real system. Overall, all successful micro-data algorithms combine these two strategies by varying the kind of model and prior knowledge. The current scientific challenges essentially revolve around scaling up to complex robots, designing generic priors, and optimizing the computing time.
\end{abstract}

\IfClass{IEEEtran}{
	\begin{IEEEkeywords}
		Learning and Adaptive Systems, Autonomous Agents, Robot Learning, Micro-Data Policy Search
	\end{IEEEkeywords}
}{}

\section{Introduction}

Reinforcement learning (RL)~\cite{sutton1998reinforcement} is a generic framework that allows robots to learn and adapt by trial-and-error. There is currently a renewed interest in RL owing to recent advances in deep learning~\cite{lecun2015deep}. For example, RL-based agents can now learn to play many of the Atari 2600 games directly from pixels~\cite{mnih2015human,mnih2016asynchronous}, that is, without explicit feature engineering, and beat the world's best players at Go and chess with minimal human knowledge~\cite{silver2017mastering}. Unfortunately, these impressive successes are difficult to transfer to robotics because the algorithms behind them are highly data-intensive: 4.8 million games were required to learn to play Go from scratch~\cite{silver2017mastering}, 38 days of play (real time) for Atari 2600 games~\cite{mnih2015human}, and, for example, about 100 hours of simulation time (much more for real time) for a 9-DOF mannequin that learns to walk~\cite{heess2017emergence}.

By contrast, robots have to face the real world, which cannot be accelerated by GPUs nor parallelized on large clusters. And the real world will not become faster in a few years, contrary to computers so far (Moore's law). In concrete terms, this means that most of the experiments that are successful in simulation cannot be replicated in the real world because they would take too much time to be technically feasible. As an example, Levine et al.~\cite{levine2018learning} recently proposed a large-scale algorithm for learning hand-eye coordination for robotic grasping using deep learning. The algorithm required approximately 800000 grasps, which were collected within a period of 2 months using 6-14 robotic manipulators running in parallel. Although the results are promising, they were only possible because they could afford having that many manipulators and because manipulators are easy to automate: it is hard to imagine doing the same with a farm of humanoids.

What is more, online adaptation is much more useful when it is fast than when it requires hours --- or worse, days --- of trial-and-error. For instance, if a robot is stranded in a nuclear plant and has to discover a new way to use its arm to open a door; or if a walking robot encounters a new kind of terrain for which it is required to alter its gait; or if a humanoid robot falls, damages its knee, and needs to learn how to limp: in most cases, adaptation has to occur in a few minutes or within a dozen trials to be of any use.

By analogy with the word ``big-data'', we refer to the challenge of learning by trial-and-error in a handful of trials as ``micro-data reinforcement learning''~\cite{mouret2016micro}. This concept is close to ``data-efficient reinforcement learning''~\cite{deisenroth_gaussian_2015}, but we think it captures a slightly different meaning. The main difference is that efficiency is a ratio between a cost and benefit, that is, data-efficiency is a ratio between a quantity of data and, for instance, the complexity of the task. In addition, efficiency is a relative term: a process is more efficient than another; it is not simply ``efficient''. In that sense, many deep learning algorithms are data-efficient because they require fewer trials than the previous generation, regardless of the fact that they might need millions of time-steps. By contrast, we propose the terminology ``micro-data learning'' to represent an absolute value, not a relative one: how can a robot learn in a few minutes of interaction? or how can a robot learn in less than 20 trials\footnote{It is challenging to put a precise limit for ``micro-data learning'' as each domain has different experimental constraints, this is why we will refer in this article to ``a few minutes'' or a ``a few trials''. The commonly used word ``big-data'' has a similar ``fuzzy'' limit that depends on the exact domain.}? Importantly, a micro-data algorithm might reduce the number of trials by incorporating appropriate prior knowledge; this does not necessarily make it more ``data-efficient'' than another algorithm that would use more trials but less prior knowledge: it simply makes them different because the two algorithms solve a different challenge.

\begin{figure*}[ht!]
	\centering
	\includegraphics[width=0.65\linewidth]{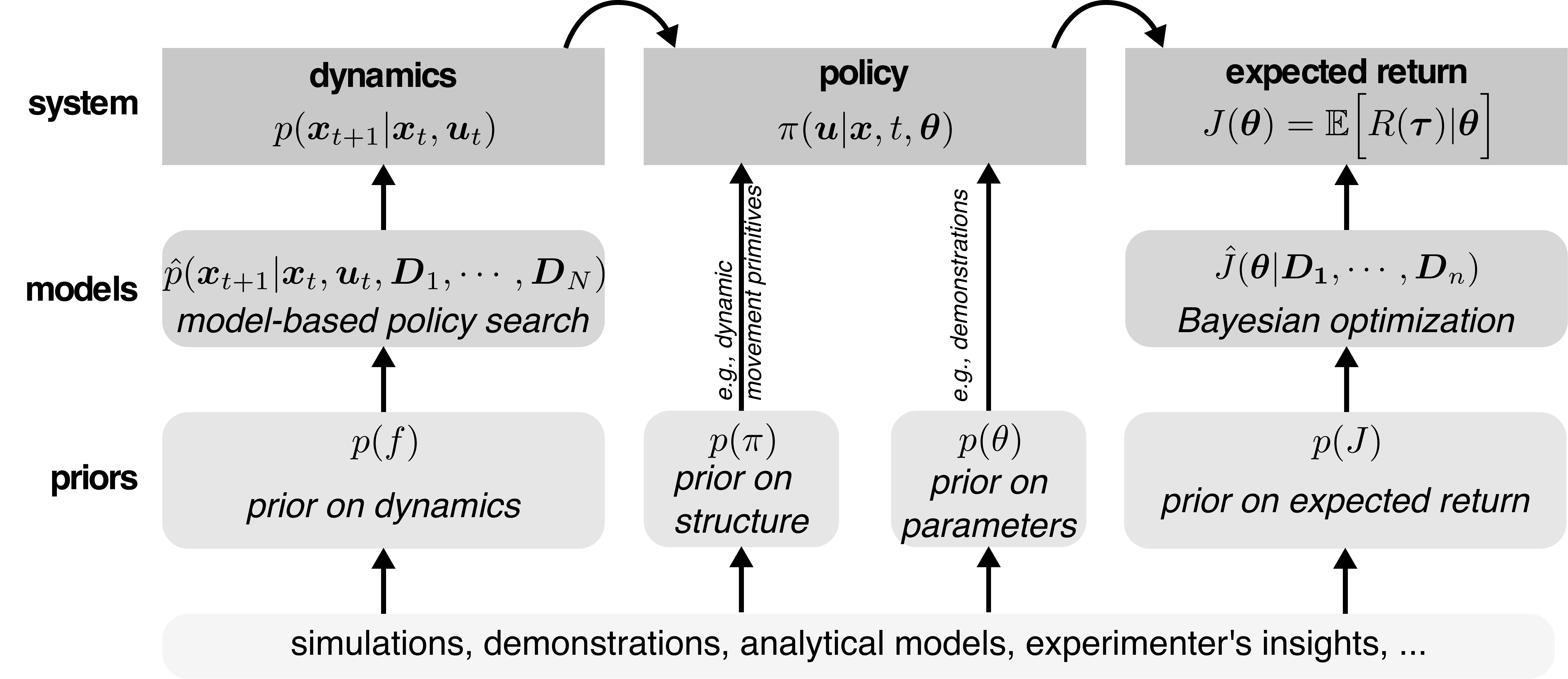}
	\caption{Overview of possible strategies for Micro-Data Policy Search (MDPS). The first strategy (bottom) is to leverage prior knowledge on the dynamics, on the policy parameters, on the structure of the policy, or on the expected return. A second strategy is to learn surrogate models of the dynamics or of the expected return.}
	\label{fig:mdps_overview}
\end{figure*}

Among the different approaches for RL, most of the recent work in robotics focuses on \emph{Policy Search} (PS), that is, on viewing the RL problem as the optimization of the parameters of a given policy~\cite{deisenroth_survey_2013} (see the problem formulation, Section~\ref{sec:problem_formulation}). A few PS algorithms are explicitly focused on requiring very little \emph{interaction time} with the robot, which often implies that they authorize themselves to substantially increase the \emph{computing time} and the amount of \emph{prior knowledge}. The purpose of this paper is to survey such existing micro-data policy search techniques that have been successfully used for robot control
\footnote{Planning-based and model-predictive control~\cite{garcia1989model} methods do not search for policy parameters, this is why they do not fit into the scope of this paper. Although trajectory-based policies and planning-based methods share the same goal, they usually search in a different space: planning algorithms search in the state-action space (e.g., joint positions/velocities), whereas policy methods will search for the optimal parameters of the policy, which can encode a subspace of the possible trajectories.},
and to identify the challenges in this emerging field. In particular, we focus on policy search approaches that have the \emph{explicit goal} of reducing the interaction time between the robot and the environment to a few seconds or minutes
\IfClass{IEEEtran}{
\footnote{The scarcity of data in robotics makes it necessary to follow specific strategies when designing learning algorithms. The authors of the present survey organized a very successful workshop on this exact topic at IROS 2017 (Micro-Data: the new frontier of robot learning?) and we think it is the right time to summarize the recent efforts in this direction:
	while there have been survey articles on policy search in the past (in particular~\cite{kober2013reinforcement,deisenroth_survey_2013}), there have been many exciting developments in the last years (e.g., 50\% of the papers cited in our survey have been published between 2013 and 2018).
	Moreover, our survey focuses on policy search algorithms that have the explicit goal of minimizing the interaction time as much as
	possible (and not RL or PS algorithms in general), whereas previous surveys had a broader region of interest. Consequently, we can be
	more thorough in our review and explain the algorithms in more detail.
}}{}.

Most published algorithms for micro-data policy search implement and sometimes combine two main strategies (Fig.~\ref{fig:mdps_overview}): leveraging prior knowledge (Sections \ref{sec:priors_policy}, \ref{sec:priors_return}, and \ref{sec:priors_dynamics}) and building surrogate models (Sections~\ref{sec:model_return} and~\ref{sec:model_based}).

Using prior knowledge requires balancing carefully between what can be realistically known before learning and what is left to be learnt. For instance, some experiments assume that demonstrations can be provided, but that they are imperfect~\cite{ng2006autonomous,kober2009learning}; some others assume that a damaged robot knows its model in its intact form, but not the damaged model~\cite{cully_robots_2015,pautrat2018bayesian,chatzilygeroudis2018using}.
This knowledge can be introduced at different places, typically in the structure of the policy (e.g., dynamic movement primitives~\cite{ijspeert2003learning}, Section~\ref{sec:priors_policy}), in the reward function (e.g., reward shaping, Section~\ref{sec:priors_return}), or in the dynamical model~\cite{abbeel2006using,chatzilygeroudis2018using} (Section~\ref{sec:priors_dynamics}).

\begin{figure}[ht!]
	\centering
	\includegraphics[width=0.9\linewidth]{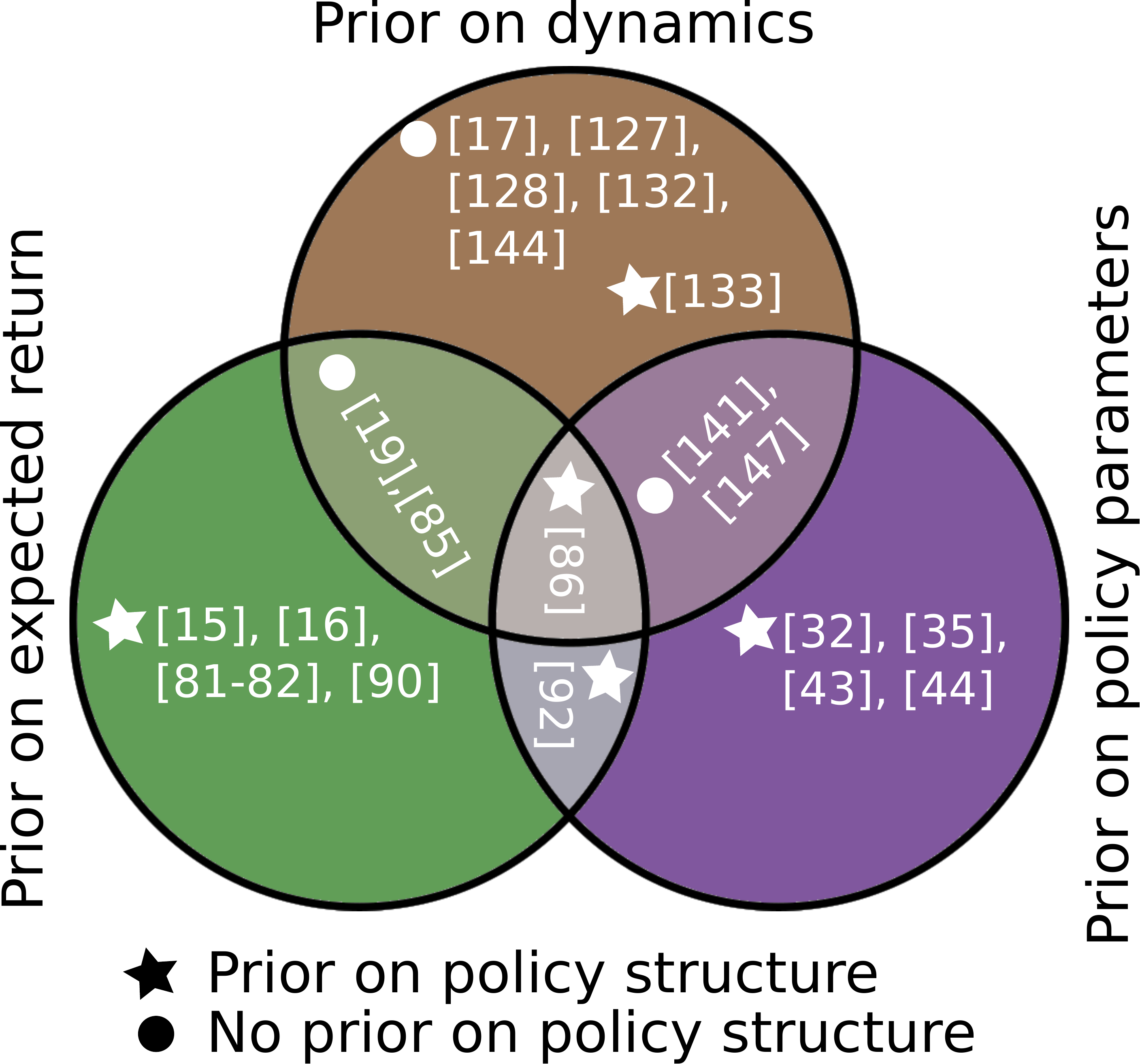}
	\caption{Main references per prior combination.}
	\label{fig:prior_refs}
\end{figure}

The second strategy is to create models from the data gathered during learning and utilize them to make better decisions about what to try next on the robot. We can further categorize these methods into (a) algorithms that learn a surrogate model of the expected return (i.e., long-term reward) from a starting state~\cite{brochu_tutorial_2010,shahriari2016taking} (Section~\ref{sec:model_return}); and (b) algorithms that learn models of the transition dynamics and/or the immediate reward function (e.g., learning a controller for inverted helicopter flight by first learning a model of the helicopter's dynamics~\cite{ng2006autonomous}, Section~\ref{sec:model_based}). The two strategies --- priors and surrogates --- are often combined (Fig.~\ref{fig:prior_refs}); for example, most works with a surrogate model impose a policy structure and some of them use prior information to shape the initial surrogate function, before acquiring any data.

This article surveys the literature along these three axes: priors on policy structure and parameters (Section~\ref{sec:priors_policy}), models of expected return (Section~\ref{sec:model_return}), and models of dynamics (Section~\ref{sec:model_based}). Section~\ref{sec:other} lists the few noteworthy approaches for micro-data policy search that do not fit well into the previous sections. Finally, Section~\ref{sec:challenges} sketches the challenges of the field and Section~\ref{sec:conclusion} proposes a few ``precepts'' and recommendations to guide future work in this field.

\section{Problem formulation} \label{sec:problem_formulation}

\noindent We model the robots as discrete-time dynamical systems that can be described by transition probabilities of the form:
\begin{align}
	\label{eq:transition_dyn}
	p(\boldsymbol{x}_{t+1}|\boldsymbol{x}_t,\boldsymbol{u}_t)
\end{align}
where the robot is at state $\boldsymbol{x}_t\in\mathbb{R}^E$ at time $t$, takes control input $\boldsymbol{u}_t\in\mathbb{R}^F$ and ends up at state $\boldsymbol{x}_{t+1}$ at time $t+1$.

If we assume deterministic dynamics and Gaussian system noise, this equation is often written as:
\begin{align}
	\label{eq:dyn}
	\boldsymbol{x}_{t+1} = f(\boldsymbol{x}_t,\boldsymbol{u}_t) + \boldsymbol{w}
	.
\end{align}
Here, $\boldsymbol{w}$ is i.i.d. Gaussian system noise, and $f$ is a function that describes the unknown transition dynamics.

We assume that the system is controlled through a parameterized \textit{policy} $\pi(\boldsymbol{u}|\boldsymbol{x},t,\boldsymbol{\theta})$ that is followed for $T$ steps ($\boldsymbol{\theta}$ are the parameters of the policy). Throughout the paper we adopt the episode-based, fixed time-horizon formulations for clarity and pedagogical reasons, but also because most of the micro-data policy search approaches use this formulation.

In the general case, $\pi(\boldsymbol{u}|\boldsymbol{x},t,\boldsymbol{\theta})$ outputs a distribution (e.g., a Gaussian) that is sampled in order to get the action to apply; i.e., we have \emph{stochastic policies}. Most algorithms utilize policies that are not time-dependent (i.e., they drop $t$), but we include it here for completeness. Several algorithms use \emph{deterministic policies}; a deterministic policy means that $\pi(\boldsymbol{u}|\boldsymbol{x},t,\boldsymbol{\theta})\Rightarrow\boldsymbol{u} = \pi(\boldsymbol{x},t|\boldsymbol{\theta})$.

When following a particular policy for $T$ time-steps from an initial state distribution $p(\boldsymbol{x}_0)$, the system's states and actions jointly form \textit{trajectories} $\boldsymbol{\tau} = (\boldsymbol{x}_0,\boldsymbol{u}_0,\boldsymbol{x}_1,\boldsymbol{u}_1,\dots,\boldsymbol{x}_T)$, which are often also called \textit{rollouts} or \textit{paths}.
We assume that a scalar performance system exists, $R(\boldsymbol{\tau})$, that evaluates the performance of the system given a trajectory $\boldsymbol{\tau}$. This \textit{long-term reward} (or \textit{return}) is defined as the sum of the immediate rewards along the trajectory $\boldsymbol{\tau}$:

\begin{align}
	\label{eq:long_term_reward}
	R(\boldsymbol{\tau}) = \sum_{t=0}^{T-1}r_{t+1} = \sum_{t=0}^{T-1}r(\boldsymbol{x}_t,\boldsymbol{u}_t,\boldsymbol{x}_{t+1})
\end{align}
where $r_{t+1} = r(\boldsymbol{x}_t,\boldsymbol{u}_t,\boldsymbol{x}_{t+1})\in\mathbb{R}$ is the \emph{immediate reward} of being in state $\boldsymbol{x}_t$ at time $t$, taking the action $\boldsymbol{u}_t$ and reaching the state $\boldsymbol{x}_{t+1}$ at time $t+1$.
We define the \textit{expected return} $J(\boldsymbol{\theta})$ as a function of the policy parameters:
\begin{align}
	\label{eq:expected_return}
	J(\boldsymbol{\theta}) & = \mathbb{E}\Big[R(\boldsymbol{\tau}) | \boldsymbol{\theta}\Big]\nonumber \\
	                       & = \int R(\boldsymbol{\tau})P(\boldsymbol{\tau} | \boldsymbol{\theta})
\end{align}
where $P(\boldsymbol{\tau}|\boldsymbol{\theta})$ is the distribution over trajectories $\boldsymbol{\tau}$ for any given policy parameters $\boldsymbol{\theta}$ applied on the actual system:
\begin{align}
	\label{eq:traj}
	\underbrace{P(\boldsymbol{\tau}|\boldsymbol{\theta})}_{\textrm{trajectories for }\boldsymbol{\theta}} = \underbrace{p(\boldsymbol{x}_0)}_\textrm{initial state} \prod_t \underbrace{p(\boldsymbol{x}_{t+1}|\boldsymbol{x}_t,\boldsymbol{u}_t)}_\textrm{transition dynamics} \underbrace{\pi(\boldsymbol{u}_t|\boldsymbol{x}_t,t,\boldsymbol{\theta})}_\textrm{policy}.
\end{align}
%

The objective of a \emph{policy search algorithm} is to find the parameters $\boldsymbol{\theta}^*$ that maximize the \textit{expected return} $J(\boldsymbol{\theta})$ when following the policy $\pi_{\boldsymbol{\theta}^*}$:
\begin{align}
	\label{eq:reward_j}
	\boldsymbol{\theta}^* = \argmax_{\boldsymbol{\theta}}J(\boldsymbol{\theta}).
\end{align}


\begin{algorithm}[tb]
	\caption{Generic policy search algorithm}
	\label{algo:generic_ps}
	\begin{algorithmic}[1]
		\label{algo:generic_ps:init}
		\State Apply initialization strategy using \textsc{InitStrategy}
		\State Collect data, $\boldsymbol{D}_0$, with \textsc{CollectStrategy}
		\For {$n=1\to N_{iter}$}
		\State Learn models using \textsc{ModelStrategy} and $\boldsymbol{D}_{n-1}$
		\State Calculate $\boldsymbol{\theta}_{n+1}$ using \textsc{UpdateStrategy}
		\State Apply policy $\pi_{\boldsymbol{\theta}_{n+1}}$ on the system
		\State Collect data, $\boldsymbol{D}_n$, with \textsc{CollectStrategy}
		\EndFor
		\State\Return $\pi_{\boldsymbol{\theta}^*} =$ \textsc{SelectBestPolicyStrategy}
	\end{algorithmic}
\end{algorithm}

Most policy search algorithms can be described with a generic algorithm (Algo.~\ref{algo:generic_ps}) and they: (1) start with an initialization strategy (\textsc{InitStrategy}), for instance using random actions, and (2) collect data from the robot (\textsc{CollectStrategy}), for instance the states at each discrete time-steps or the reward at the end of the episode; they then (3) enter a loop (for $N_{iter}$ iterations) that alternates between learning one or more models (\textsc{ModelStrategy}) with the data acquired so far, and selecting the next policy $\pi_{\boldsymbol{\theta}_{n+1}}$ to try on the robot (\textsc{UpdateStrategy}). Finally, they return the ``optimal'' policy using \textsc{SelectBestPolicyStrategy}.

This generic outline allows us to describe direct (e.g., policy gradient algorithms~\cite{sutton2000policy}), surrogate-based (e.g., Bayesian optimization~\cite{brochu_tutorial_2010}) and model-based policy search algorithms, where each algorithm implements in a different way each of \textsc{InitStrategy}, \textsc{CollectStrategy}, \textsc{ModelStrategy} and \textsc{UpdateStrategy}. We will also see that in this outline we can also fit policy search algorithms that utilize priors; coming from simulators, demonstrations or any other source.

To better understand how policy search is performed, let us use a gradient-free optimizer (\textsc{UpdateStrategy}) and learn directly on the system (i.e., \textsc{ModelStrategy} $= \emptyset$). This type of algorithm falls in the category of \emph{model-free} or \emph{direct} policy search algorithms~\cite{sutton1998reinforcement,kohl_policy_2004}.
\textsc{InitStrategy} can be defined as randomly choosing some policy parameters, $\boldsymbol{\theta}_1$ (Algo.~\ref{algo:model_free_direct}), and \textsc{CollectStrategy} collects samples of the form $(\boldsymbol{\theta}, \frac{\sum_i^N R(\boldsymbol{\tau})_i}{N})$ by running $N$ times the policy $\pi_{\boldsymbol{\theta}}$. We execute the same policy multiple times because we are interested in approximating the expected return (Eq.~\eqref{eq:long_term_reward}). $\tilde{J}_{\boldsymbol{\theta}} = \frac{\sum_i^N R(\boldsymbol{\tau})_i}{N}$ is then used as the value for the sample $\boldsymbol{\theta}$ in a regular optimization loop that tries to maximize it (i.e., the \textsc{UpdateStrategy} is optimizer-dependent).

\begin{algorithm}[tb]
	\caption{Gradient-free direct policy search algorithm}
	\label{algo:model_free_direct}
	\begin{algorithmic}[1]
		\Procedure{\textsc{InitStrategy}}{}
		\label{algo:model_free_direct:init}
		\State Select $\boldsymbol{\theta}_1$ randomly
		\EndProcedure
		\Procedure{\textsc{CollectStrategy}}{}
		\label{algo:model_free_direct:collect}
		\State Collect samples of the form $(\boldsymbol{\theta}, \frac{\sum_i^N R(\boldsymbol{\tau})_i}{N}) = (\boldsymbol{\theta}, \tilde{J}_{\boldsymbol{\theta}})$ by running policy $\pi_{\boldsymbol{\theta}}$ $N$ times.
		\EndProcedure
	\end{algorithmic}
\end{algorithm}

This straightforward approach to policy search typically requires a large amount of interaction time with the system to find a high-performing solution~\cite{sutton1998reinforcement}. Many approaches have been suggested to improve the sample efficiency of model-free approaches (e.g.,~\cite{silver2014deterministic,mnih2016asynchronous,degris2012linear,ciosek2018expectedPG,van2009theoretical,sutton2000policy,schulman2015trust,lillicrap2016continuous,abdolmaleki2015model}). Nevertheless, the objective of the present article is to describe algorithms that require several orders of magnitude less interaction time by leveraging priors and models.


\section{Using priors on the policy parameters/representation} \label{sec:priors_policy}

\newcommand{\task}{\ensuremath{\zeta}\xspace}

When designing the policy
$\pi(\boldsymbol{u}|\boldsymbol{x},t,\boldsymbol{\theta})$, the key design choices are what the space of $\boldsymbol{\theta}$ is, and how it maps states to actions.
This design
is guided by a trade-off between having a representation that is \emph{expressive}, and one that provides a space that is \emph{efficiently searchable}.

Expressiveness can be defined in terms of the optimal policy $\pi^*_\task$.
For a given task $\task$, there is theoretically always at least one optimal policy $\pi^*_\task$. Here, we drop $\boldsymbol{\theta}$ to express that we do not mean a specific representation parameterized by $\boldsymbol{\theta}$. Rather $\pi^*_\task$ emphasizes that there is some policy (with some representation, perhaps unknown to us) that cannot be outperformed by any other policy (whatever its representation). We use $J_\task(\pi^*_\task)$ to denote this highest possible expected reward.

A parameterized policy $\pi_{\boldsymbol{\theta}}$ should be expressive enough to {\em represent} this optimal policy $\pi^*_\task$ (or at least come close), i.e.,

\begin{align}
	J_\task(\pi^*_\task) - \max_{\boldsymbol{\theta}}J_\task(\boldsymbol{\theta})<\delta
\end{align}

where $\delta$ is some acceptable margin of suboptimality. Note that absolute optimality is rarely required in robotics; in many everyday applications, small tracking errors may be acceptable, and the quadratic command cost does not need to be at the absolute minimum.

On the other hand, the policy representation should be such that it is easy (or at least feasible) to find $\boldsymbol{\theta}^*$, i.e., it should be {\em efficiently searchable}\footnote{Analogously, the universal approximation theorem states that a feedforward network with single hidden layer suffices to {\em represent} any continuous function, but it does not imply that the function is {\em learnable} from data.}. In general, smaller values of $\dim(\boldsymbol{\theta})$ lead to more efficiently searchable spaces.

In the following subsections, we describe several common policy representations, which make different trade-offs between expressiveness and being efficiently searchable, and several common strategies to improve the generality and convergence of policy search algorithms.

\subsection{Hand-designed policies}

One approach to reducing the policy parameter space is to hand-tailor it to the task $\task$ to be solved. In~\cite{fidelman2004learning}, for instance, a policy for ball acquisition is designed. The resulting policy only has only four parameters, i.e., $\dim(\boldsymbol{\theta})$ is 4. This low-dimensional policy parameter space is easily searched, and only 672 trials are required to optimize the policy.
Thus, prior knowledge is used to find a compact representation, and policy search is used to find the optimal $\boldsymbol{\theta}^*$ for this representation.

One disadvantage of limiting $\dim(\boldsymbol{\theta})$ to a very low dimensionality is that $\delta$ may become quite large, and we have no estimate of how much more the reward could have been optimized with a more expressive policy representation. Another disadvantage is that the representation is very specific to the task \task for which it was designed. Thus, such a policy cannot be reused to learn other tasks.
It then greatly limits the transfer learning capabilities of the approaches, since the learned policy can hardly be re-used for any other task.

\subsection{Policies as function approximators}

Ideally, our policy representation $\Theta$ is expressive enough so that we can apply it to many different tasks, i.e.,
\begin{align}
	\argmin_{\Theta} \sum_{n=1}^N J_{\task_n}(\pi^*_{\task_n}) - \max_{\boldsymbol{\theta}}J_{\task_n}(\boldsymbol{\theta})\mbox{, with $\boldsymbol{\theta} \in \Theta$},
\end{align}
i.e., over a set of tasks, we minimize the sum of differences between the theoretically optimal policy $\pi^*$ for each task, and the optimal policy \emph{given the representation} $\pi_{\boldsymbol{\theta}}$ for each task\footnote{Note that this optimization is never actually performed. It is a mathematical description of what the policy representation designer is implicitly aiming for.}.
A few examples of such generally applicable policy representations are
linear policies, radial basis function networks, and neural networks (NN).
%
These more general policies can be used for many tasks~\cite{guenter2007reinforcement,kober2013reinforcement}. However, prior knowledge is still required to determine the appropriate number of basis functions and their shape.
Non-parametric methods partially alleviate the need to such these parameters~\cite{hoof2015learning}, but the number of basis functions (one for each data point) may become very large and slow down learning.
Again, a lower number of basis functions will usually lead to more efficient learning, but less expressive policies and thus potentially higher $\delta$.

One advantage of using a function approximator is that demonstrations can often be used to determine the initial policy parameters. The initial parameters $\boldsymbol{\theta}_1$ can be obtained through supervised learning or other machine learning techniques, by providing the demonstration as training data $(\boldsymbol{x}_i,\boldsymbol{u}_i)_{i=1:N}$. This is discussed in more detail in Section~\ref{sec:imitation}.

The function approximator can be used to generate a single estimate (corresponding to a first order moment in statistics), but it can also be extended to higher order moments. Typically, extending it to second order moments allows the system to get information about the variations that we can exploit to fulfill a task, as well as the synergies between the different policy parameters in the form of covariances. This is typically more expensive to learn---or it requires multiple demonstrations \cite{matsubara2011learning}---but the learned representation can typically be more expressive, facilitating adaptation and generalization.

\subsection{Trajectory-based policies}

\newcommand{\vtheta}{\ensuremath{\mathbf{\boldsymbol{\theta}}}}
\newcommand{\timeconstant}{\ensuremath{\omega}}
\newcommand{\taskspace}{\ensuremath{\boldsymbol{\xi}}}

Trajectory-based policy types have been widely used in the robot learning literature~\cite{khansari2011learning,ude2016trajectory,ude2010task,spitz2017trial,stulp2013robot}, and especially within the policy search problem for robotics~\cite{ijspeert13dynamical,ijspeert02movement,stulp2013robot}. This type of policy is well-suited for several typical classes of tasks in robotics, such as point-to-point movements or repetitive movements. There exist basically two types of trajectory-based policies:
(1)~way-point based policies~\cite{roy2002motion}, and
(2)~dynamical system based~\cite{khansari2011learning,ijspeert02movement}.

One approach to encoding trajectories is to define the policy as a sequence of way-points. In~\cite{roy2002motion}, the authors define the problem of motion planning as a policy search problem where the parameters of the policy are the concatenated way-points, $\boldsymbol{w}_i$. They were able to define an algorithm that outperforms several baselines including dynamic programming.

Policies based on dynamical systems have been used more extensively within the robot learning literature as they combine the generality of function approximators with the advantages of dynamical systems, such as robustness towards perturbations and stability guarantees~\cite{ijspeert13dynamical,ijspeert02movement,khansari2011learning,stulp2013robot}, which are desirable properties of a robotic system.

Perhaps the most widely used trajectory-based policy type within the policy search framework is Dynamical Movement Primitives (DMPs); we can categorize them into discrete DMPs and rhythmic DMPs depending on the type of motion they are describing (point-to-point or repetitive).


Discrete DMPs are summarized in Eq.~\eqref{equ_dmp}. The canonical system represents the movement \emph{phase} $s$, which starts at 1, and converges to 0 over time. The transformation systems combines a spring-damper system with a function approximator $f_\vtheta$, which, when integrated, generates accelerations $\ddot{\taskspace}$. Multi-dimensional DMPs are achieved by coupling multiple transformation systems with one canonical system. The vector $\taskspace$ typically represents the end-effector pose or the joint angles.

As the spring-damper system converges to $\taskspace^g$, and $s$ (and thus $s\, f_\vtheta(s)$) converges to 0, the overall system $\taskspace$ is guaranteed to converge to $\taskspace^g$. We have:
\begin{align}
	\timeconstant \ddot{\taskspace} & = \underbrace{\alpha(\beta(\taskspace^g-\taskspace)-\dot{\taskspace})}_{\mbox{\footnotesize Spring-damper system}} ~~+~~ \underbrace{s\, f_\vtheta(s)}_{\mbox{\footnotesize Forcing term}}, & \mbox{\footnotesize (Transf.)}\label{equ_dmp}          \\
	\timeconstant \dot{s}           & = -\alpha_s s.                                                                                                                                                                              & \mbox{\footnotesize (Canonical)} \label{equ_canonical}
\end{align}

This facilitates learning, because, whatever parameterization $\vtheta$ of the function approximator we choose, a discrete DMP is guaranteed to converge towards a goal $\taskspace^g$.
Similarly, a rhythmic DMP will always generate a repetitive motion, independent of the values in $\boldsymbol{\theta}$. The movement can be made slower or faster by changing the time constant \timeconstant.

Another advantage of DMPs is that only one function approximator is learned for each dimension of the DMP, and that the input of each function approximator is the phase variable $s$, which is always 1D. Thus, whereas the overall DMP closes the loop on the state $\taskspace$, the part of the DMP that is learned ($f_\vtheta(s)$) is an open-loop system. This greatly facilitates learning, and simple black-box optimization algorithms have been shown to outperform state-of-the-art RL algorithms for such policies~\cite{stulp13policy}. Approaches for learning the goal $\taskspace^g$ of a discrete movement have also been proposed~\cite{stulp12reinforcement}. Since the goal is constant throughout the movement, few trials are required to learn it.


The optimal parameters $\boldsymbol{\theta}^*$ for a certain DMP are specific to one specific task \task. Task-parameterized (dynamical) motion primitives aim at generalizing them to variations of a task, which are described with the task parameter vector $\boldsymbol{q}$ (e.g., the 3D pose to place an object on a table~\cite{stulp13learning} or the 3D pose of the end-effector~\cite{ude2010task}).
Similar approaches can be used in contextual policies, see e.g.,~\cite{kupcsik2017model,abdolmaleki2017contextual}.
Learning a motion primitive that is optimal for all variations of a task (i.e., all $\boldsymbol{q}$ within a range) is much more challenging, because the curse of dimensionality applies to the task parameter vector $\boldsymbol{q}$ just as it does for the state vector $\boldsymbol{x}$ in reinforcement learning. 
Task-parameterized representations based on the use of multiple coordinate systems have been developed to cope with this curse of dimensionality~\cite{Calinon16JIST}. These models have only been applied to learning from demonstration applications so far.



DMPs, nevertheless, are time-dependent and thus can produce behaviors that are not desirable; for example, a policy that cannot adapt to perturbations after some time. Stable Estimator of Dynamical Systems (SEDS)~\cite{khansari2011learning} explores how to use dynamical systems in order to define autonomous (i.e., time-independent) controllers (or policies) that are asymptotically stable. The main idea of the algorithm is to use a finite mixture of Gaussian functions as the policy, $\dot{\boldsymbol{\xi}} = \pi_{\text{seds}}(\boldsymbol{\xi})$, with specific properties that satisfy some stability guarantees. SEDS, however, requires demonstrated data in order to optimize the policy (i.e., data gathered from experts), although similar ideas have been used within the RL framework~\cite{guenter2007reinforcement}.

It is important to note that if $\taskspace$ or $\boldsymbol{w}$ are not defined in joint space (i.e., the control variables), then most of the approaches assume the existence of a low-level controller that can take target accelerations, velocities or positions (in $\taskspace$ or $\boldsymbol{w}$) and produce the appropriate low-level control commands (e.g., torques) to achieve these targets. Moreover, all the stability and convergence guarantees mentioned in this section apply solely on the behavior or policy dynamics (e.g., stability or convergence of the desired velocity profile in the end-effector space) and not on the robotic system as a whole\footnote{One would need to analyze the complete system of the policy, low-level controllers, and robot dynamics to see if the whole system behavior is stable.}.

%



\subsection{Learning the controller}


If the policy generates a reference trajectory, a controller is required to map this trajectory (and the current state) to robot control commands (typically torques or joint angle velocity commands). This can be done for instance with a \emph{proportional-integral-derivative} (PID) controller~\cite{buchli11learning}, or a \emph{linear quadratic tracking} (LQT) controller~\cite{Calinon14ICRA}. The parameters of this controller can also be included in $\boldsymbol{\theta}$, so that both the reference trajectory and controller parameters are learned at the same time. By doing so, appropriate gains~\cite{buchli11learning,Calinon13RAS} or forces~\cite{kalakrishnan11learning} for the task can be learned together with the movement required to reproduce the task. Typically, such representation provides a way to coordinate motor commands to react to perturbations, by rejecting perturbations only in the directions that would affect task performance.


\subsection{Learning the policy representation}

So far we have described how the policy representation is determined with prior knowledge, and the $\boldsymbol{\theta}$ of this policy is then optimized through policy search. Another approach is to learn the policy representation and its parameters at the same time, as in NeuroEvolution of Augmenting Topologies (NEAT)~\cite{stanley2002evolving}. It is even possible, in simulation, to co-evolve an appropriate body morphology and policy~\cite{sims1994evolving,bongard2003evolving}. These approaches, however, require massive amounts of rollouts, and do not focus on learning in a handful of trials.

\subsection{Hierarchical and Symbolic Policy Representations}

To further generalize policies to different contexts, several approaches have been proposed. Daniel et al. propose the use of a hierarchical policy composed of a gating network and multiple sub-policies, and introducing an entropy-based constraint ensuring that the agent finds distinct solutions with different sub-policies~\cite{daniel2016hierarchical}. These sub-policies are treated as latent variables in an expectation-maximization procedure, allowing the distribution of the update information between the sub-policies.
Higher layers of the hierarchy may be replaced with symbolic representations, as in~\cite{ryan98rltops,lang2012exploration,yang2018peorl}. A full discussion of the many approaches in this area is beyond the scope of this article.


\subsection{Initialization with demonstrations / imitation learning}\label{sec:imitation}

An advantage of using expressive policies is that they are able to learn (close to) optimal policies for many different tasks. A downside is that such policies are also able to represent many suboptimal policies for a particular task, i.e., there will be many local minima.
To ensure convergence, it is important that the initial policy parameters are close to the global optimum. In robotics, this is possible through imitation~\cite{osa2018algorithmic,argall2009survey,billard2008robot}, i.e., the initialization of $\boldsymbol{\theta}$ from a demonstrated trajectory.
Starting with a $\boldsymbol{\theta}$ that is close $\boldsymbol{\theta}^*$ greatly reduces the number of samples to find $\boldsymbol{\theta}^*$, and the interplay between imitation and policy search is therefore an important component in micro-data learning.

Initialization with demonstrations is possible if we know the general movement a robot should make to solve the task, and if we can demonstrate it, either by recording our movement, by teleoperating the robot, or by physically guiding the robot through kinesthetic teaching.
Each of these modalities has some limitations. Observational learning does not take into account differences between user and robot (in terms of embodiment, kinematic and dynamic capabilities). Dynamic or skillful tasks are difficult to demonstrate by teleoperation and kinesthetic teaching. Recording both force and position information is limited with kinesthetic teaching and observational learning.

\begin{messagebox}{msg:imitation_trajectory}
	Using policy structures that are inspired or derived by prior knowledge about the task or the robot at hand is an effective way of creating a policy representation that is expressive enough but also efficiently searchable. If it is further combined with learning from demonstrations (or imitation learning), then it can lead to powerful approaches that are able to learn in just a handful of trials.

	\emph{Recommended readings:} \cite{osa2018algorithmic,billard2008robot}
\end{messagebox}

\section{Learning models of the expected return} \label{sec:model_return}

With the appropriate policy representation (and/or initial policy parameters) chosen, the policy search in Algorithm 1 is then executed. The most important step is determining the next parameter vector $\boldsymbol{\theta}_{n+1}$ to test on the physical robot.

In order to choose the next parameter vector $\boldsymbol{\theta}_{n+1}$ to test on the physical robot, a strategy is to learn a model $\hat{J}(\boldsymbol{\theta})$ of the expected return $J(\boldsymbol{\theta})$ (Eq.~\eqref{eq:expected_return}) using the values collected during the previous episodes, and then choose the optimal $\boldsymbol{\theta}_{n+1}$ according to this model. Put differently, the main concept is to optimize $J(\boldsymbol{\theta})$ by leveraging $\hat{J}(\boldsymbol{\theta} | R(\boldsymbol{\tau} | \boldsymbol{\theta}_1), \cdots, R(\boldsymbol{\tau} |\boldsymbol{\theta}_N))$.

\subsection{Bayesian optimization: active learning of policy parameters} \label{sec:bayesian_optimization}

\begin{algorithm}[htb]
	\caption{Policy search with Bayesian optimization}
	\label{algo:bo_ps}
	\begin{algorithmic}[1]
		\Procedure{\textsc{CollectStrategy}}{}
		\State Collect samples of the form $(\boldsymbol{\theta}, R(\boldsymbol{\tau}))$
		\EndProcedure
		\Procedure{\textsc{ModelStrategy}}{}
		\State Learn model $\hat{J}:\boldsymbol{\theta}\to J(\boldsymbol{\theta})$
		\EndProcedure
		\Procedure{\textsc{UpdateStrategy}}{}
		\label{algo:bo_ps:opt}
		\State $\boldsymbol{\theta}_{n+1} = \argmax_{\boldsymbol{\theta}}\textsc{AcquisitionFunction}(\boldsymbol{\theta}|\hat{J})$
		\EndProcedure
	\end{algorithmic}
\end{algorithm}

The most representative class of algorithms that falls in this category is Bayesian optimization (BO)~\cite{brochu_tutorial_2010}. BO consists of two main components: a model of the \emph{expected return}, and an \emph{acquisition function}, which uses the model to define the utility of each point in the search space.

BO for policy search follows the generic policy search algorithm (Algo.~\ref{algo:generic_ps}) and implements \textsc{CollectStrategy}, \textsc{ModelStrategy} and \textsc{UpdateStrategy} (Algo.~\ref{algo:bo_ps}). More specifically, a surrogate model, $\hat{J}(\boldsymbol{\theta})$, of the expected return is learned from the data, then the next policy to test is selected by optimizing the \textsc{AcquisitionFunction}. The \textsc{AcquisitionFunction} tries to intelligently exploit the model and its uncertainties in order to trade-off exploration and exploitation.

The main axes of variation are: (a) the way \textsc{InitStrategy} is defined (the most usual approaches are random policy parameters or random actions), (b) the type of model used to learn $J$, (c) which \textsc{AcquisitionFunction} is used, and (d) the optimizer used to optimize the \textsc{AcquisitionFunction}.

\para{Gaussian Processes} Gaussian Process (GP) regression~\cite{rasmussen2006gaussian} is the most popular choice for the model. A GP is an extension of multivariate Gaussian distribution to an infinite-dimension stochastic process for which any finite combination of dimensions will be a Gaussian distribution~\cite{rasmussen2006gaussian}. More precisely, it is a distribution over functions, completely specified by its mean function, $m(\cdot)$ and covariance function, $k(\cdot,\cdot)$ and it is computed as follows:
\begin{align}
	\hat{J}(\boldsymbol{\theta})\sim\mathcal{GP}(m(\boldsymbol{\theta}),k(\boldsymbol{\theta},\boldsymbol{\theta}')).
\end{align}
Assuming $D_{1:t} = \{R(\boldsymbol{\tau}|\boldsymbol{\theta}_1),...,R(\boldsymbol{\tau}|\boldsymbol{\theta}_t)\}$ is a set of observations, we can query the GP at a new input point $\boldsymbol{\theta}_*$ as follows:
\begin{align}
	\label{eq:gp}
	p(\hat{J}(\boldsymbol{\theta}_*)|D_{1:t},\boldsymbol{\theta}_*) = \mathcal{N}(\mu(\boldsymbol{\theta}_*),\sigma^{2}(\boldsymbol{\theta}_*)).
\end{align}
The mean and variance predictions of the GP are computed using a kernel vector $\pmb{k} = k(D_{1:t},\boldsymbol{\theta}_*)$, and a kernel matrix $K$, with entries $K_{ij} = k(\boldsymbol{\theta}_i,\boldsymbol{\theta}_j)$:
\begin{align}
	\label{eq:gp_detail}
	\mu(\boldsymbol{\theta}_*)        & = \pmb{k}^{T}K^{-1}D_{1:t}\nonumber\mbox{,}                                \\
	\sigma^{2}(\boldsymbol{\theta}_*) & = k(\boldsymbol{\theta}_*,\boldsymbol{\theta}_*)-\pmb{k}^{T}K^{-1}\pmb{k}.
\end{align}

For the acquisition function, most algorithms use the Expected Improvement, the Upper Confidence Bound or the Probability of Improvement~\cite{brochu_tutorial_2010,hennig2012entropy}.

\para{Probability of Improvement} One of the first acquisition functions is the Probability of Improvement~\cite{kushner1964new} (PI). PI defines the probability that a new test point $\hat{J}(\boldsymbol{\theta})$ will be better than the best observation so far $\boldsymbol{\theta}^+$; since we cannot directly get this information from $D_{1:t}$, in practice we query the approximated model $\hat{J}$ on $D_{1:t}$ and get the best parameters. When using GPs as the surrogate model, this can be analytically computed:

\begin{align}
	\label{eq:pi}
	PI(\boldsymbol{\theta}) & = p(\hat{J}(\boldsymbol{\theta}) > \hat{J}(\boldsymbol{\theta}^+))\nonumber                                   \\
	                        & = \Phi\Big(\frac{\mu(\boldsymbol{\theta}) - \hat{J}(\boldsymbol{\theta}^+)}{\sigma(\boldsymbol{\theta})}\Big)
\end{align}
where $\Phi(\cdot)$ denotes the CDF of the standard normal distribution.
The main drawback of PI is that it basically performs pure exploitation; in practice, a slightly modified version of PI is used where a trade-off parameter $\xi$ is added~\cite{brochu_tutorial_2010}.

\para{Expected Improvement} The Expected Improvement~\cite{brochu_tutorial_2010} (EI) acquisition function is an extension of PI, where the expected improvement (deviation) from the current maximum is calculated. Again, when using GPs as the surrogate model, EI can be analytically computed:

\begin{align}
	\label{eq:ei}
	 & I(\boldsymbol{\theta}) = \text{max}\{0, \hat{J}(\boldsymbol{\theta}) - \hat{J}(\boldsymbol{\theta}^+)\}\nonumber \\
	 & EI(\boldsymbol{\theta}) = \mathbb{E}(I(\boldsymbol{\theta}))\nonumber                                            \\
	 & = \begin{cases}
		(\mu(\boldsymbol{\theta}) - \hat{J}(\boldsymbol{\theta}^+))\Phi(Z) + \sigma(\boldsymbol{\theta})\phi(Z), & \text{if $\sigma(\boldsymbol{\theta})>0$}. \\
		0,                                                                                                       & \text{otherwise}.
	\end{cases}                                                                                 \\
	 & Z = \frac{\mu(\boldsymbol{\theta}) - \hat{J}(\boldsymbol{\theta}^+)}{\sigma(\boldsymbol{\theta})}\nonumber
\end{align}
where $\phi(\cdot)$ and $\Phi(\cdot)$ denote the PDF and CDF of the standard normal distribution respectively.

\para{Upper Confidence Bound} The Upper Confidence Bound (UCB) acquisition function is the easiest to grasp and works very well in practice~\cite{hennig2012entropy}. When using GPs as the surrogate model, it is defined as follows:

\begin{align}
	\label{eq:ucb}
	 & UCB(\boldsymbol{\theta}) = \mu(\boldsymbol{\theta}) + \alpha\sigma(\boldsymbol{\theta})
\end{align}
where $\alpha$ is a user specified parameter. When using UCB as the acquisition function, it might be difficult to choose $\alpha$ and the initial hyper-parameters of the kernel (that affect $\sigma$) as the range of $J$ and $\boldsymbol{\theta}$ plays a huge role on this. The GP-UCB algorithm~\cite{srinivas2009gaussian,brochu_tutorial_2010} automatically adjusts $\alpha$ and provides some theoretical guarantees on the regret bounds of the algorithm.

\para{Entropy Search} The Entropy Search (ES)~\cite{hennig2012entropy} acquisition function selects policy parameters in order to maximally reduce the uncertainty about the location of the maximum of $J(\boldsymbol{\theta})$ in each step. It quantifies this uncertainty through the entropy of the distribution over the location of the maximum,  $p_{\text{max}}(\boldsymbol{\theta}) = \mathbb{P}(\boldsymbol{\theta}\in\argmin_{\boldsymbol{\theta}}J(\boldsymbol{\theta}))$. ES basically defines a different \textsc{AcquisitionFunction} for BO as follows:

\begin{align}
	\label{eq:entropy_search_acqui}
	ES(\boldsymbol{\theta}) = \argmax_{\boldsymbol{\theta}}\mathbb{E}[\Delta H(\boldsymbol{\theta})]
\end{align}

\noindent where $\Delta H(\boldsymbol{\theta})$  is the change in entropy of $p_{\text{max}}$ caused by retrieving a new cost value at location $\boldsymbol{\theta}$.

A thorough experimental analysis~\cite{hennig2012entropy} concluded that EI can perform better than PI and UCB on artificial objective functions, but more recent experiments on gait learning on a physical robot suggested that UCB can outperform EI in real situations~\cite{calandra2015bayesian}. In most cases, ES outperforms all other acquisition functions at a bigger computation cost~\cite{hennig2012entropy}.

Martinez-Cantin et al.~\cite{martinez2007active} were among the first to use BO as a policy search algorithm; in particular, their approach was able to learn a policy composed of way-points in order to control a mobile robot that had to navigate in an uncertain environment. Since BO is not modeling the dynamics of the system/robot, it can be effective for learning policies for robots with complex (e.g., locomotion tasks, because of the non-linearity created by the contacts) or high-dimensional dynamics. For instance, Bayesian optimization was successfully used to learn policies for a quadruped robot~\cite{Lizotte2007} (around 100 trials with a well-chosen 15D policy space), a small biped ``compass robot''~\cite{calandra2015bayesian} (around 100 trials with a finite state automata policy), and a pocket-sized, vibrating soft tensegrity robot~\cite{rieffel2017soft} (around 30 trials with directly controlling the motors). In all of these cases, BO was at least an order of magnitude more data-efficient than competing methods.

Unfortunately, BO scales badly with respect to the dimensionality of the policy space because modeling the objective function (i.e., the expected return) becomes exponentially harder when the dimension increases~\cite{bellman1957dynamic}. This is why all the aforementioned studies employed low-dimensional policy spaces and very well chosen policy structures (i.e., they all use a strong prior on the policy structure).
Scaling up BO is, however, an active field of research and various promising approaches (e.g., random embeddings~\cite{wang2016bayesian} and additive models~\cite{kandasamy2015high,rolland2018high,akrour2017local}) could be applied to robotics in the future. Combining stochastic optimization with learned local models of the expected return can be an alternative to BO and could scale much better with respect to the policy dimensions~\cite{abdolmaleki2015model}.

\subsection{Bayesian optimization with priors: using non-zero mean functions as a starting point for the search process} \label{sec:priors_return}

One of the most interesting features of BO is that it can leverage priors (e.g., from simulation or from previous tasks) to accelerate learning on the actual task. Perhaps the most representative algorithm in this area is  the ``Intelligent Trial \& Error'' (IT\&E) algorithm~\cite{cully_robots_2015}. IT\&E first uses MAP-Elites~\cite{cully_robots_2015}, an evolutionary illumination~\cite{mouret2015illuminating, vassiliades2017cvt} (also known as quality-diversity~\cite{pugh2016quality}) algorithm, to create a repertoire of about 15000 high-performing policies and stores them in a low-dimensional map (e.g., 6-dimensional whereas the policy space is 36-dimensional). When the robot needs to adapt, a BO algorithm searches for the best policy in the low-dimensional map and uses the reward stored in the map as the mean function of a GP. This algorithm allowed a 6-legged walking robot to adapt to several damage conditions (e.g., a missing or a shortened leg) in less than 2 minutes (less than a dozen of trials), whereas it used a simulator of the intact robot to generate the prior.

\para{Gaussian processes with priors}{Assuming $D_{1:t} = \{R(\boldsymbol{\tau}|\boldsymbol{\theta}_1),\allowbreak...,R(\boldsymbol{\tau}|\boldsymbol{\theta}_t)\}$ is a set of observations and $R_m(\boldsymbol{\theta})$ being the reward in the map, we can query the GP at a new input point $\boldsymbol{\theta}_*$ as follows:
	\begin{align}
		\label{eq:gp_j_priors}
		p(\hat{J}(\boldsymbol{\theta}_*)|D_{1:t},\boldsymbol{\theta}_*) = \mathcal{N}(\mu(\boldsymbol{\theta}_*),\sigma^{2}(\boldsymbol{\theta}_*)).
	\end{align}
	The mean and variance predictions of this GP are computed using a kernel vector $\pmb{k} = k(D_{1:t},\boldsymbol{\theta}_*)$, and a kernel matrix $K$, with entries $K^{ij} = k(\boldsymbol{\theta}_i,\boldsymbol{\theta}_j)$ and where $k(\cdot,\cdot)$ is the kernel of the GP:
	\begin{align}
		\label{eq:gp_j_priors_detail}
		\mu(\boldsymbol{\theta}_*)        & = R_m(\boldsymbol{\theta}_*)+\pmb{k}^{T}K^{-1}(D_{1:t}-R_m(\boldsymbol{\theta}_{1:t}))\mbox{,}\nonumber \\
		\sigma^{2}(\boldsymbol{\theta}_*) & = k(\boldsymbol{\theta}_*,\boldsymbol{\theta}_*)-\pmb{k}^{T}K^{-1}\pmb{k}.
	\end{align}
	The formulation above allows us to combine observations from the prior and the real-world smoothly. In areas where real-world data is available, the prior's prediction will be corrected to match the real-world ones. On the contrary, in areas far from real-world data, the predictions resort to the prior function~\cite{cully_robots_2015,lee2017gp,chatzilygeroudis2018reset}.}

Following a similar line of thought but implemented differently, a few recent works~\cite{antonova2016sample,antonova2017deep} use a simulator to learn the kernel function of a GP, instead of utilizing it to create a mean function like in IT\&E~\cite{cully_robots_2015}. In particular, Antonova et al.~\cite{antonova2016sample} used domain knowledge for bipedal robots (i.e., \emph{Determinants of Gait} (DoG)~\cite{inman1953major}) to produce a kernel that encodes the differences in walking gaits rather than the Euclidean distance of the policy parameters. In short, for each controller parameter $\boldsymbol{\theta}$ a score $\text{sc}(\boldsymbol{\theta})$ is computed by summing the 5 DoG and the kernel $k(\cdot,\cdot)$ is defined as $k(\boldsymbol{\theta}_i,\boldsymbol{\theta}_j) = k(\text{sc}(\boldsymbol{\theta}_i), \text{sc}(\boldsymbol{\theta}_j))$. This approach outperformed both traditional BO and state-of-the-art black-box optimizers (Covariance Matrix Adaptation Evolution Strategies; CMA-ES~\cite{hansen2001completely}). 
Moreover, in a follow-up work~\cite{antonova2017deep}, the authors use NNs to model this kernel instead of hand-specifying it. Their evaluation shows that the learned kernels perform almost as good as hand-tuned ones and outperform traditional BO. Lastly, in this work they were able to make a physical humanoid robot (ATRIAS) walk in a handful of trials.

A similar but more general idea (i.e., no real assumption about the underlying system) was introduced by~\cite{wilson_using_2014}. The authors propose a Behavior-Based Kernel (BBK) that utilizes trajectory data to compare policies, instead of using the distance in parameters (as is usually done). More specifically, they define the behavior of a policy to be the associated trajectory density $P(\boldsymbol{\tau} | \boldsymbol{\theta})$ and the kernel $k(\cdot,\cdot)$ is defined as $k(\boldsymbol{\theta}_i,\boldsymbol{\theta}_j) = \exp{(-\alpha\cdot D(\boldsymbol{\theta}_i,\boldsymbol{\theta}_j))}$, where $D(\boldsymbol{\theta}_i,\boldsymbol{\theta}_j)$ is defined as a sum of KL-divergences between the trajectory densities of different policies. Their approach was able to efficiently learn on several benchmarks; e.g., it required on average less than 20 episodes on the mountain car, acrobot and cartpole swing-up tasks. One could argue that this approach does not utilize any prior information, but rather creates it on the fly; nevertheless, the evaluation was only performed with low-dimensional and well-chosen policy spaces.

Wilson et al.~\cite{wilson_using_2014} proposed to learn models of the dynamics and the immediate reward to compute an approximate mean function of the GP, which is then used in a traditional BO procedure. They also combine this idea with the BBK kernel and follow a regular BO procedure where at each iteration they re-compute the mean function of the GP with the newly learned models. Although, their approach successfully learned several tasks in less than 10 episodes (e.g., mountain car, cartpole swing-up), there is an issue that might not be visible at first sight: the authors combine model learning, which scales badly with the state/action space dimensionality (see Section~\ref{sec:model_based}), with Bayesian optimization, which scales badly with the dimensionality of the policy space. As such, their approach can only work with relatively small state/action spaces and small policy spaces. Using priors on the dynamics (see Section~\ref{sec:priors_dynamics}) and recent improvements on BO (see Section~\ref{sec:bayesian_optimization}) could make their approach more practical.

Lober et al.~\cite{lober2016efficient} use a BO procedure that selects parameterizations of a QP-based whole body controller~\cite{salini2011synthesis,spitz2017trial} in order to control a humanoid robot. In particular, they formulate a policy that includes the QP-based controller (that contains a model of the system and an optimizer) and is parameterized by way-points (and/or switching times). Their approach was able to allow an iCub robot to move a heavy object while maintaining body balance and avoid collisions~\cite{lober2016efficient,lober2017optimizing}.

\para{Multiple information sources}{
Instead of using the simulator to precompute priors, Alonso et al.~\cite{marco17virtualvsreal} propose an approach that has the ability to automatically decide whether it will gain crucial information from a real sample or it can use the simulator that is cheaper. More specifically, they present a BO algorithm for multiple information sources. Their approach relies on entropy search (see Eq.~\eqref{eq:entropy_search_acqui}) and they use entropy to measure the information content of simulations and real experiments. Since this is an appropriate unit of measure for the utility of both sources, the algorithm is able to compare physically meaningful quantities in the same units, and trade off accuracy for cost. As a result, the algorithm can automatically decide whether to evaluate cheap, but inaccurate simulations or perform expensive and precise real experiments. They applied their method, called \emph{Multifidelity Entropy Search} (MF-ES), to fine-tune the policy of a cart-pole system and showed that their approach can speed up the optimization process significantly compared to standard BO.

Pautrat et al.~\cite{pautrat2018bayesian} also recently proposed to combine BO with multiple information sources (or \emph{priors}). They define a new \textsc{AcquisitionFunction} function for BO, which they call \emph{Most Likely Expected Improvement} (MLEI). MLEI attempts to have the right balance between the likelihood of the priors and the potential for high-performing solutions. In other words, a good expected improvement according to an unlikely model should be ignored; conversely, a likely model with a low expected improvement might be too pessimistic (``nothing works'') and not helpful. A model that is ``likely enough'' and lets us expect some good improvement might be the most helpful to find the maximum of the objective function. The MLEI acquisition function is defined as follows:
\begin{align}
	\label{eq:mlei_acqui}
	EIP(\boldsymbol{\theta}, \mathcal{P}) = EI(\boldsymbol{\theta}) \times p(\hat{J}(\boldsymbol{\theta}_{1..t}) ~|~ \boldsymbol{\theta}_{1..t}, \boldsymbol{\mathcal{P}}(\boldsymbol{\theta}_{1..t}))\nonumber \\
	MLEI(\boldsymbol{\theta}, \mathcal{P}_1, \cdots, \mathcal{P}_m) = \max_{\boldsymbol{p} \in \mathcal{P}_1, \cdots, \mathcal{P}_m} EIP(\boldsymbol{\theta}, \boldsymbol{p})
\end{align}
\noindent where $\mathcal{P}_i, i=1\dots m$ is the set of available priors (where each $\mathcal{P}_i$ is defined similarly to $R_m$ in Eq.\eqref{eq:gp_j_priors_detail}). They evaluated their approach in a transfer learning scenario with a simulated arm and in a damage recovery one with both a simulated and a physical hexapod robot. Their approach demonstrates improved performance relative to random trials or a hand-chosen prior (when that prior does not correspond to the new task). Interestingly, this method also is able to outperform the real prior in some circumstances.
}

\para{Safety-Aware Approaches}{
	Another interesting direction of research is using variants of BO for safety-aware learning; that is learning that actively tries to avoid regions that might cause harm to the robot. In~\cite{papaspyros2016safety} the authors proposed an extension of IT\&E that safely trades-off between exploration and exploitation in a damage recovery scenario. To achieve this, (1) they generate, through MAP-Elites, a diverse archive of estimations concerning performance and safety criteria and (2) they use this as prior knowledge in a constrained BO~\cite{gardner2014bayesian} procedure that guides the search towards a compensatory behavior and with respect to the safety beliefs. Their algorithm, sIT\&E, allowed a simulated damaged iCub to crawl again safely.

	Similarly, in~\cite{berkenkamp16safe} Berkenkamp et al. introduced SafeOpt, a BO procedure to automatically tune controller parameters by trading-off between exploration and exploitation only within a safe zone of the search space. Their approach requires minimal knowledge, such as an initial, not optimal, safe controller to bootstrap the search.
	This allowed a quadrotor vehicle to safely improve its performance over the initial policy.

}


\begin{messagebox}{msg:bayesian_optimization}
	Bayesian optimization is an active learning framework for micro-data reinforcement learning that is effective when using uncertainty-based models and when there exists some prior on the structure of the policy or on the expected return. However, BO is limited to low-dimensional policy spaces.

	\emph{Recommended readings:} \cite{Lizotte2007,cully_robots_2015}
\end{messagebox}
\section{Learning models of the dynamics} \label{sec:model_based}
Instead of learning a model of the expected long-term reward (Section~\ref{sec:bayesian_optimization}), one can also learn a model of the dynamics of the robot. By repeatedly querying this surrogate model, it is then possible to make a prediction of the expected return. This idea leads to \emph{model-based policy search algorithms}~\cite{deisenroth_survey_2013,polydoros2017survey}, in which the trajectory data are used to learn the dynamics model, then policy search is performed on the model~\cite{sutton1991dyna,kaelbling1996reinforcement}.

Put differently, the algorithms leverage the trajectories $\boldsymbol{\tau}_1, \cdots, \boldsymbol{\tau}_N$ observed so far to learn a function $\hat{f}(\boldsymbol{x}, \boldsymbol{u})$ such that:
\begin{equation}
	\hat{\boldsymbol{x}}_{t+1} = \hat{f}(\boldsymbol{x}_t, \boldsymbol{u}_t).
\end{equation}
This function, $\hat{f}(\boldsymbol{x}_t, \boldsymbol{u}_t)$, is then used to compute an estimation of the expected return, $\hat{J}(\boldsymbol{\theta}| \boldsymbol{\tau}_1, \cdots, \boldsymbol{\tau}_N)$.

\subsection{Model-based Policy Search: alternating between updating the model and learning a policy in the model} \label{sec:model_based_ps}
Let us consider that the actual dynamics $f$ (and consequently the transition probabilities) are approximated by a model $\hat{f}$ and the immediate reward function $r$ is approximated by a model $\hat{r}$. As such, in model-based policy search we are alternating between learning the models ($\hat{f}$ and $\hat{r}$) and maximizing the expected long-term reward on the model:
\begin{equation}
	\label{eq:reward_j_model}
	\hat{J}(\boldsymbol{\theta}) = \mathbb{E}[\hat{R}(\boldsymbol{\tau}) | \boldsymbol{\theta}] = \int \hat{R}(\boldsymbol{\tau})\hat{P}(\boldsymbol{\tau} | \boldsymbol{\theta})
\end{equation}
where
\begin{equation}
	\label{eq:traj_model}
	\hat{P}(\boldsymbol{\tau} | \boldsymbol{\theta}) = p(\boldsymbol{x}_0)\prod_t \hat{p}(\boldsymbol{x}_{t+1}|\boldsymbol{x}_t,\boldsymbol{u}_t)\pi_{\boldsymbol{\theta}}(\boldsymbol{u}_t|\boldsymbol{x}_t,t).
\end{equation}
%
\vspace{-0.8em}
\begin{equation}
	\label{eq:long_term_reward_model}
	\hat{R}(\boldsymbol{\tau}) = \sum_{t=0}^{T-1}\hat{r}_{t+1} = \sum_{t=0}^{T-1}\hat{r}(\boldsymbol{x}_t,\boldsymbol{u}_t,\boldsymbol{x}_{t+1})
\end{equation}

This iterative scheme can be seen as follows:
\begin{align}
	\boldsymbol{\tau}_n       & \sim P(\boldsymbol{\tau} | \boldsymbol{\theta}_n)                            \\
	\boldsymbol{D}_n          & = \boldsymbol{D}_{n-1}\cup\{\boldsymbol{\tau}_n,R(\boldsymbol{\tau}_n)\}     \\
	\boldsymbol{\theta}_{n+1} & = \argmax_{\boldsymbol{\theta}}\hat{J}(\boldsymbol{\theta}|\boldsymbol{D}_n)
\end{align}
where $\boldsymbol{\theta}_0$ is randomly determined or initialized to some value, $\boldsymbol{D}_0 = \emptyset$ and $\hat{J}(\boldsymbol{\theta}|\boldsymbol{D})$ means calculating $\hat{J}(\boldsymbol{\theta})$ once the models $\hat{f}$ and $\hat{r}$ are learned using the dataset of trajectories and rewards $\boldsymbol{D}$. 



\begin{algorithm}[htb]
	\caption{Model-based policy search}
	\label{algo:mb_ps}
	\begin{algorithmic}[1]
		\Procedure{\textsc{CollectStrategy}}{}
		\State Collect samples of the form $(\boldsymbol{x}_t,\boldsymbol{u}_t, r_{t+1})$
		\EndProcedure
		\Procedure{\textsc{ModelStrategy}}{}
		\State Learn model $\hat{f} : (\boldsymbol{x}_t,\boldsymbol{u}_t)\to\boldsymbol{x}_{t+1}$
		\State Learn model $\hat{r} : (\boldsymbol{x}_t,\boldsymbol{u}_t,\boldsymbol{x}_{t+1})\to r_{t+1}$
		\EndProcedure
		\Procedure{\textsc{UpdateStrategy}}{}
		\label{algo:mb_ps:opt}
		\State $\boldsymbol{\theta}_{n+1} = \argmax_{\boldsymbol{\theta}}\hat{J}(\boldsymbol{\theta}|\boldsymbol{D}_n)$
		\EndProcedure
	\end{algorithmic}
\end{algorithm}

Model-based policy search follows the generic policy search algorithm (Algo.~\ref{algo:generic_ps}) and implements \textsc{CollectStrategy}, \textsc{ModelStrategy} and \textsc{UpdateStrategy} (Algo.~\ref{algo:mb_ps}). The main axes of variation are: (a) the way \textsc{InitStrategy} is defined (the most usual approaches are random policy parameters or random actions), (b) the type of models used to learn $\hat{f}$ and $\hat{r}$, (c) the optimizer used to optimize $\hat{J}(\boldsymbol{\theta}|\boldsymbol{D}_n)$, and (d) how are the long-term predictions, given the models, performed (i.e., how Eq.~\eqref{eq:reward_j_model} is calculated or approximated).

Model-based policy search algorithms are usually more data-efficient than both direct and surrogate-based policy search methods as they do not depend much on the dimensionality of the policy space. On the other hand, since they are modeling the transition dynamics, practical algorithms are available only for relative small state-action spaces~\cite{deisenroth_survey_2013,polydoros2017survey}.

\subsubsection{Model learning}
\label{sec:model_learning}
There exist many approaches to learn the models $\hat{f}$ and $\hat{r}$ (for model-based policy search) in the literature~\cite{tangkaratt2014model,levine2014learning,deisenroth_gaussian_2015}. Most algorithms assume a known reward function; otherwise they usually use the same technique to learn both models. We can categorize the learned models in deterministic (e.g., NNs or linear regression) and probabilistic ones (e.g., GPs).

Probabilistic models usually rely on Bayesian methods and are typically non-parametric, whereas deterministic models are typically parametric. Probabilistic models are usually more effective than deterministic models in model-based policy search~\cite{deisenroth_survey_2013,parmas18pipps} because they provide uncertainty information that can be incorporated into the long-term predictions, thus giving the capability to the optimizer to find more robust controllers (and not over-exploit the model biases). Black-DROPS~\cite{chatzilygeroudis2017black} and PILCO~\cite{deisenroth2011pilco} both utilize GPs to greatly reduce the interaction time to solve several tasks, although Black-DROPS is not tied to them and any deterministic or probabilistic model can be used.

The model-based Policy Gradients with Parameter-based Exploration algorithm~\cite{tangkaratt2014model} suggested to directly estimate the transition probabilities $p(\boldsymbol{x}_{t+1}|\boldsymbol{x}_t,\boldsymbol{u}_t)$ using least-squares conditional density estimation~\cite{sugiyama2010least}, instead of learning the model $\hat{f}$. This formulation allowed to bypass some drawbacks of GPs such as computation speed and smoothness assumption (although choosing appropriate kernels in the GPs can produce non-smooth predictions).

Another way of learning models of the dynamics is to use local linear models~\cite{levine2016end,levine2014learning,kumar2016optimal}; i.e., models that are trained on and are only correct in the regions where one controller/policy can drive the system. Guided policy search with unknown dynamics utilizes this scheme and is able to learn efficiently even in high-dimensional states and discontinuous dynamics, like 2D walking and peg-in-the-hole tasks~\cite{levine2016end,levine2014learning} and even dexterous manipulation tasks~\cite{kumar2016optimal}.

There has, also, recently been some work on using Bayesian NNs (BNNs)~\cite{gal2016dropout} to improve the scaling of model-based policy search algorithms~\cite{gal2016improving,higuera2018synthesizing}. Compared to GPs, BNNs scale much better with the number of samples. Nevertheless, BNNs require more tedious hyper-parameter optimization and there is no established, intuitive way to include prior knowledge (apart from the structure). A combination of ensembles and probabilistic NNs has been recently proposed~\cite{chua2018deeprl} for learning probabilistic dynamics models of higher dimensional systems; for example, state-of-the-art performance was obtained in controlling the half-cheetah benchmark~\cite{wawrzynski2007learning} by combining these models with model-predictive control. Recent works showcase that using BNNs with stochastic inputs (and the appropriate policy search procedure) is beneficial when learning in scenarios with multi-modality and heteroskedasticity~\cite{depeweg2017learning}; traditional model learning approaches (e.g., GPs) fail to properly model these scenarios. Moreover, decomposing aleatoric (i.e., inherent uncertainty of the underlying system) and epistemic (i.e., uncertainty due to limited data) uncertainties in BNNs (with latent input variables) can provide useful information on which points to sample next~\cite{depeweg2018decomposition}.

Lastly, when performing model-based policy search under partial observability, different model learning techniques should be used. One interesting idea is to optimize the model with the explicit goal of explaining the already observed trajectories instead of focusing on the step-by-step predictions. Doerr et al.~\cite{doerr2017optimizing} recently proposed a principled approach to incorporate these ideas into GP modeling and were able to outperform other robust models in long-term predictions and showcase improved performance for model-based policy search on a real robot with noise and latencies.

\subsubsection{Long-term predictions}

Traditionally, we would categorize the model-based policy search algorithms in those that perform \textit{stochastic long-term predictions} by means of samplings and those that perform \textit{deterministic long-term predictions} by deterministic inference techniques~\cite{deisenroth_survey_2013}. Recently, an alternative way of computing the expected long-term reward was introduced by~\cite{chatzilygeroudis2017black} (\textit{Policy Evaluation as a Noisy Observation}), where the trajectory generation is combined with the optimization process in order to achieve high-quality predictions with fewer Monte-Carlo rollouts.

\paragraph{Stochastic long-term predictions}
\label{sec:stoch_long}

The actual dynamics of the system are approximated by the model $\hat{f}$, and the immediate reward function by the model $\hat{r}$. The model $\hat{f}$ provides the transition probabilities $\hat{p}(\boldsymbol{x}_{t+1}|\boldsymbol{x}_t,\boldsymbol{u}_t)$. Similarly, the model $\hat{r}$ provides the immediate reward $\hat{r}_{t+1} = \hat{r}(\boldsymbol{x}_t,\boldsymbol{u}_t,\boldsymbol{x}_{t+1})$. When applying a policy (with some parameters $\boldsymbol{\theta}$) on the model, we get a \textit{rollout} or \textit{trajectory}:
\begin{align}
	\boldsymbol{\tau} & = (\boldsymbol{x}_0,\boldsymbol{u}_0,\boldsymbol{x}_1,\boldsymbol{u}_1,\dots,\boldsymbol{x}_T) \\
	\boldsymbol{r}    & = (\hat{r}_1,\hat{r}_2,\dots,\hat{r}_T)                                                        
\end{align}
where
\begin{align}
	\boldsymbol{x}_0     & \sim p(\boldsymbol{x}_0)                                                           \\
	\hat{r}_{t+1} &= \hat{r}(\boldsymbol{x}_t,\boldsymbol{u}_t,\boldsymbol{x}_{t+1})\\
	\boldsymbol{u}_t     & \sim \pi_{\boldsymbol{\theta}}(\boldsymbol{u}_t|\boldsymbol{x}_t,t)                \\
	\boldsymbol{x}_{t+1} & \sim \hat{p}(\boldsymbol{x}_{t+1}|\boldsymbol{x}_t,\boldsymbol{u}_t).
\end{align}
This is basically sampling the distribution over trajectories, $\hat{P}(\boldsymbol{\tau} | \boldsymbol{\theta})$, which is feasible since the sampling is performed with the models. When applying the same policy (i.e., a policy with the same parameters $\boldsymbol{\theta}$), the trajectories $\boldsymbol{\tau}$ (and consequently $\boldsymbol{r}$) can be different (i.e., stochastic) because (of at least one of the following):
\begin{itemize}
	\item The policy is stochastic. If the policy is deterministic, then $\boldsymbol{u}_t = \pi_{\boldsymbol{\theta}}(\boldsymbol{x}_t,t)$;
	\item The models ($\hat{f}$ and/or $\hat{r}$) are probabilistic;
	\item Of the initial state distribution, $p(\boldsymbol{x}_0)$.
\end{itemize}

\emph{Monte-Carlo \& PEGASUS policy evaluation:}
Once we know how to generate trajectories given some policy parameters, we need to define the way to evaluate the performance of these policy parameters. Perhaps the most straightforward way of computing the expected log-term reward of some policy parameters is to generate $m$ trajectories with the same policy along with their long-term costs and then compute the average (i.e., perform Monte-Carlo sampling):
\begin{align}
	\tilde{\hat{J}}(\boldsymbol{\theta}) = \frac{1}{m}\sum_{i=1}^m{\hat{R}_i(\boldsymbol{\tau}^i)}.
\end{align}
A more efficient way of computing the expected long-term reward with stochastic trajectories is with the PEGASUS sampling procedure~\cite{ng2000pegasus}. In the PEGASUS sampling procedure the random seeds for each time step are fixed. As a result, repeating the same experiment (i.e., the same sequence of control inputs and the same initial state) would result into exactly the same trajectories. This significantly reduces the sampling variance compared to pure Monte-Carlo sampling and can be shown that optimizing this \textit{semi-stochastic} version of the model is equivalent to optimizing the actual model.

The advantages of the sampling-based policy evaluations schemes are that each \textit{rollout} can be performed in parallel and that they require much less implementation effort than the deterministic long-term predictions (see Section~\ref{sec:det_long}). Nevertheless, these sampling-based procedures can experience big variances in the predictions that may negatively affect the optimization process.
In~\cite{kupcsik2017model} the authors showed that when using enough sample trajectories, better approximations of the expected return can be obtained than the ones of deterministic long-term predictions (see Section~\ref{sec:det_long}); moreover, computation time can be greatly reduced by exploiting the parallelization capabilities of modern GPUs. Another recent work~\cite{chua2018deeprl} also strongly justifies the use of sampling-based policy evaluations over deterministic inference methods (especially in higher dimensional systems).

%
\emph{Probabilistic Inference for Particle-based Policy Search (PIPPS):}
Recently, Parmas et al.~\cite{parmas18pipps} proposed the PIPPS algorithm which effectively combines the Reparameterization gradients (RP) and the Likelihood ratio gradients (LR); they call them Total Propagation (TP). Their paper showcases that LR gradients (and their combined TP gradients) do not suffer from the curse of chaos (or exploding gradients), whereas RP gradients require a very large number of rollouts to accurately estimate the gradients, even for simple problems.

\paragraph{Deterministic long-term predictions}
\label{sec:det_long}
Instead of sampling trajectories $\boldsymbol{\tau}$, the probability distribution $\hat{P}(\boldsymbol{\tau} | \boldsymbol{\theta})$ can be computed with deterministic approximations, such as linearization~\cite{anderson1979optimal}, sigma-point methods~\cite{julier2004unscented} or moment matching~\cite{deisenroth_gaussian_2015}. All these inference methods attempt to approximate the original distribution with a Gaussian.

Assuming a joint probability distribution $\hat{p}(\boldsymbol{x}_t,\boldsymbol{u}_t) = \mathcal{N}(\boldsymbol{\mu}_t,\boldsymbol{\Sigma}_t)$, the distribution $\hat{P}(\boldsymbol{\tau} | \boldsymbol{\theta})$ can be computed by successively computing the distribution of $\hat{p}(\boldsymbol{x}_{t+1})$ given $\hat{p}(\boldsymbol{x}_t,\boldsymbol{u}_t)$. Computing $\hat{p}(\boldsymbol{x}_{t+1})$ corresponds to solving the integral:
\begin{align}
	\hat{p}(\boldsymbol{x}_{t+1}) = \iiint\hat{p}(\boldsymbol{x}_{t+1}|\boldsymbol{x}_t,\boldsymbol{u}_t)\hat{p}(\boldsymbol{x}_t,\boldsymbol{u}_t)d\boldsymbol{x}_t d\boldsymbol{u}_t d\boldsymbol{w}.
\end{align}
This integral can be computed analytically only if the transition dynamics $\hat{f}$ are linear (in that case $\hat{p}(\boldsymbol{x}_{t+1})$ is Gaussian). This is rarely the case and as such, approximate inference techniques are used. Usually, we approximate $\hat{p}(\boldsymbol{x}_{t+1})$ as a Gaussian; this can be done either by linearization~\cite{anderson1979optimal}, sigma-point methods~\cite{julier2004unscented} or moment matching~\cite{deisenroth_gaussian_2015}. The PILCO algorithm~\cite{deisenroth2011pilco} uses moment matching, which is the best unimodal approximation of the predictive distribution in the sense that it minimizes the 
KL-divergence between the true predictive distribution and the unimodal approximation~\cite{deisenroth_survey_2013}.

One big advantage of using deterministic inference techniques for long-term predictions is the low-variance they exhibit in the predictions. In addition, using these inference techniques allows for analytic gradient computation and as such we can exploit efficient gradient-based optimization. However, each of these inference techniques has its own disadvantages; for example, exact moments (for moment matching) can be computed only in special cases since the required integrals might be intractable, which limits the overall approach (e.g., PILCO requires that the reward function is known and differentiable).

The PILCO algorithm~\cite{deisenroth_gaussian_2015} uses this type of long-term predictions and it was the first algorithm that showed remarkable data-efficiency on several benchmark tasks (e.g., less than 20 seconds of interaction time to solve the cart-pole swing-up task)~\cite{deisenroth2011pilco}. It was also able to learn on a physical low-cost manipulator~\cite{deisenroth_learning_2011} and simulated walking tasks~\cite{deisenroth2012toward} among the many successful applications of the algorithm~\cite{deisenroth_gaussian_2015}.

\paragraph{Policy evaluation as a noisy observation}
This approach~\cite{chatzilygeroudis2017black} exploits the \textit{implicit averaging} property~\cite{jin2005evolutionary,miller1996genetic,tsutsui1997genetic} of population, rank-based optimizers, like CMA-ES~\cite{hansen2006cma}, in order to perform sampling-based evaluation of the trajectories efficiently (i.e., reducing the computation time of the policy search on the model).
%
The key idea is that when using this type of optimizers, the problem can be transformed into a noisy optimization one, thus, there is no need to (fully) compute the expected long-term reward, as this expectation can be implicitly computed by the optimizer. Similar ideas have been previously explored for model-free policy search~\cite{heidrich2009hoeffding}.

In more detail, instead of performing deterministic long-term predictions, like PILCO, or Monte-Carlo evaluation, like PEGASUS, Black-DROPS stochastically generates trajectories, but considers that each of these trajectories (or rollouts) is a measurement of a function $G(\boldsymbol{\theta})$ that is the actual function $\hat{J}(\boldsymbol{\theta})$ perturbed by a noise $N(\boldsymbol{\theta})$:
\begin{align}
	\label{eq:objective_g}
	G(\boldsymbol{\theta}) & = \hat{J}(\boldsymbol{\theta})  + N(\boldsymbol{\theta}).
\end{align}

It is easy to verify that maximizing $\mathbb{E}\Big[G(\boldsymbol{\theta})\Big]$ is equivalent to maximizing $\hat{J}(\boldsymbol{\theta})$, when $\mathbb{E}\Big[N(\boldsymbol{\theta})\Big] = \text{constant}$.

\emph{Implicit averaging and noisy functions:}
Seeing the maximization of $\hat{J}(\boldsymbol{\theta})$ as the optimization of a noisy function allows to maximize it without computing or estimating it explicitly.
The Black-DROPS algorithm ulitizes a recent variant of CMA-ES (i.e., one of the most successful algorithms for optimizing noisy and black-box functions~\cite{jin2005evolutionary,hansen2009method,hansen2009benchmarking_noisy}) that 
combines random perturbations with re-evaluation for uncertainty handling~\cite{hansen2009method} along with restart strategies for better exploration~\cite{auger2005restart}.

While Black-DROPS has the same data-efficiency as PILCO, it has the added benefit of being able to exploit multi-core architectures, thus, greatly reducing the computation time~\cite{chatzilygeroudis2017black}. Similar to most Monte-Carlo methods (like GP-REPS~\cite{kupcsik2017model}), Black-DROPS is a purely black-box model-based policy search algorithm; i.e., one can swap the model types, reward functions and/or initialization procedure with minimal effort. This is an important feature as it allows us to more easily exploit good sources of prior information~\cite{chatzilygeroudis2018using}. Black-DROPS was able to learn in less than 20 seconds of interaction time to solve the cartpole swing-up task as well as to control a physical 4-DOF physical manipulator in less than 5-6 episodes.

\subsection{Using priors on the dynamics} \label{sec:priors_dynamics}

Reducing the interaction time in model-based policy search can be achieved by using priors on the models~\cite{bischoff_policy_2014,deisenroth_multi_task_2014,chatzilygeroudis2018using,cutler2015efficient,lee2017gp,saveriano2017data,wu2012semi}; i.e., starting with an initial guess of the dynamics (and/or the reward function) and then learning the residual model. This type of algorithm follow the general model-based policy search framework (Algo.~\ref{algo:mb_ps}) and usually implement different types of \textsc{InitStrategy}.
Notably, the most successful approaches rely on GPs to model the dynamics, as priors can be very elegantly incorporated.

\para{Gaussian processes with priors for dynamical models}{Assuming $D_{1:t} = \{f(\boldsymbol{\tilde{x}}_1),...,f(\boldsymbol{\tilde{x}}_t)\}$ is a set of observations, $\boldsymbol{\tilde{x}}_t = (\boldsymbol{x}_t,\boldsymbol{u}_t)\in\mathbb{R}^{E+F}$ and $M(\boldsymbol{\tilde{x}})$ being the simulator function (i.e., the initial guess of the dynamics), we can query the GP at a new input point $\boldsymbol{\tilde{x}}_*$ similar to Eq.~\eqref{eq:gp_j_priors}-\eqref{eq:gp_j_priors_detail} (we provide only the mean prediction for notation):
\begin{align}
	\label{eq:gp_priors_detail}
	\mu(\boldsymbol{\tilde{x}}_*) & = M(\boldsymbol{\tilde{x}}_*)+\pmb{k}^{T}K^{-1}(D_{1:t}-M(\boldsymbol{\tilde{x}}_{1:t}))
\end{align}
Of course, we have $E$ independent GPs; one for each output dimension~\cite{chatzilygeroudis2017black,deisenroth2011pilco}.

A few approaches~\cite{ko2007gaussian,bischoff_policy_2014} use simple analytic and fast simulators to create a GP prior of the dynamics (and assume the reward function to be known). PILCO with priors~\cite{cutler2015efficient} uses simulated data (from running PILCO in the simulator) to create a GP prior for the dynamics and then performs policy search with PILCO. It was able to increase the data-efficiency of PILCO in a real inverted pendulum using a very simple model as a prior. A similar approach, PI-REM~\cite{saveriano2017data}, utilizes analytic equations for the dynamics prior and tries to actively bring the real trials as close as possible to the simulated ones (i.e., reference trajectory) using a slightly modified PILCO policy search procedure. PI-REM was also able to increase the data-efficiency of PILCO in a real inverted pendulum (with variable stiffness actuators) using a simple model as a prior.

%

Black-DROPS with priors~\cite{chatzilygeroudis2018using} proposes a new GP learning scheme that combines model identification and non-parametric model learning (called GP-MI) and then performs policy search with Black-DROPS. The main idea of GP-MI is to use simulators with tunable parameters, i.e., mean functions of the form $M(\boldsymbol{\tilde{x}}, \boldsymbol{\phi}_M)$ where each vector $\boldsymbol{\phi}_M\in\mathbb{R}^{n_M}$ corresponds to a different prior model of the system (e.g., different lengths of links). Searching for the $\boldsymbol{\phi}_M$ that best matches the observations can be seen as a model identification procedure, which could be solved via minimizing the mean squared error; nevertheless, the authors formulate it in a way so that they can exploit the GP framework to jointly optimize for the kernel hyper-parameters and the mean parameters, which allows the modeling procedure to balance between non-parametric and parametric modeling.

Black-DROPS with GP-MI was able to robustly learn controllers for a pendubot swing-up task~\cite{spong1995pendubot} even when the priors were misleading. More precisely, it was able to outperform Black-DROPS, PILCO, PILCO with priors, Black-DROPS with fixed priors (i.e., this should be similar to PI-REM) 
and IT\&E. Moreover, Black-DROPS with GP-MI was able to find high-performing walking policies for a physical damaged hexapod robot (48D state and 18D action space) in less than 1 minute of interaction time and outperformed IT\&E that excels in this setting~\cite{chatzilygeroudis2018using,cully_robots_2015}.

Following a similar rationale, VGMI~\cite{zhu2018fast}, uses a Bayesian optimization procedure to find the simulator's mechanical parameters so as to match the real-world trajectories (i.e., it performs model identification) and then performs policy search on the updated simulator. In particular, VGMI was able to learn policies for a physical dual-arm collaborative task and out-performed PILCO.

Finally, an approach that splits the self-modeling process from the policy search is presented in~\cite{bongard2006resilient}. The authors were among the first ones to combine a self-modeling procedure (close to model identification~\cite{siciliano2016springer}) with policy search. The self-modeling part of their approach consists of 3 steps: (a) action executing and data-collection, (b) synthesization of 15 candidate self-models that explain the sensory data and (c) active selection of the action that will elicit the most information from the robot. After a few cycles of these steps (i.e., around 15), the most accurate model is selected and policy search is performed to produce a desired behavior. Their approach was able to control in less than 20 episodes a four-legged robot and it was also able to adapt to damages in a few trials (by re-running the self-modeling procedure).



\begin{messagebox}{msg:mbps_priors}
	Model-based policy search algorithms are the most data-efficient algorithms, especially when they take into account the uncertainty of the model. While they typically suffer from the curse of dimensionality (state/action space), endowing them with prior knowledge on the dynamics can reduce their interaction time requirements even when learning with high-dimensional or complicated systems. The main challenge in this direction is to overcome the computational complexity of the approaches.

	\emph{Recommended readings:} \cite{deisenroth_gaussian_2015,chatzilygeroudis2017black, chatzilygeroudis2018using}
\end{messagebox}

\section{Other approaches}\label{sec:other}

\subsection{Guided policy search}

Guided policy search (GPS) with unknown dynamics~\cite{levine2016end,levine2014learning} is a somewhat hybrid approach that combines local trajectory optimization (that happens directly on the real system), learning local models of the dynamics (see Section~\ref{sec:model_learning}) and indirect policy search where it attempts to approximate the local controllers with one big NN policy (using supervised learning). In more detail, GPS consists of two loops: an outer loop that executes the local linear-Gaussian policies on the real system, records data and fits the dynamics models and an inner loop where it alternates between optimizing the local linear-Gaussian policies (using trajectory optimization and the fitted dynamics models) and optimizing the global policy to match all the local policies (via supervised learning and without utilizing the learned models)~\cite{levine2016end}.

The results of GPS show that it is less data-efficient than model-based policy search approaches, but more data-efficient than traditional direct policy search. Moreover, GPS is able to handle bigger state-action spaces (i.e., it has also been used with image observations~\cite{levine2016end}) than traditional model-based policy search approaches as it reduces the final policy optimization step in a supervised one that can be efficiently tackled with all the recent deep learning methods~\cite{lecun2015deep}. GPS was able to learn in less than 100 episodes even in high-dimensional states and discontinuous dynamics like 2D walking, peg-in-the-hole task and controlling an octopus robot~\cite{levine2016end,levine2014learning} among the many successful applications of the algorithm~\cite{montgomery2017reset,levine2013guided}.

\subsection{Transferability approaches}
The main hypothesis of the transferability approach~\cite{koos2013transferability,koos2013fast} is that physics simulators are accurate for some policies, e.g., static gaits, and inaccurate for some others, e.g., highly dynamic gaits.  As a consequence, it is possible to learn in simulation if the search is constrained to policies that are simulated accurately.
As no simulator currently comes with an estimate of its accuracy, the key idea of the transferability approach is to learn a model of a \emph{transferability function}, which predicts the accuracy of a simulator given policy parameters or a trajectory in simulation. This function is often easier to learn than the expected return because this is essentially a
classification problem (instead of regression). In addition, small errors in the model have often little consequences, because the search is mainly driven by the expected return in simulation (and not by the transferability optimization).

The resulting learning process requires only a handful trials on the physical robot (in most of the experiments, less than 25); however, the main drawback is that it can only find policies that perform similarly in simulation and in reality (e.g., static gaits versus highly dynamic gaits). These type of algorithms were able to efficiently learn policies for mobile robots that have to navigate in mazes~\cite{koos2013transferability} (15 trials on the robot), for a walking quadruped robot~\cite{koos2013transferability,koos2012online} (about 10 trials) and for a 6-legged robot that had to learn how to walk in spite of a damaged leg without updating the simulator~\cite{koos2013fast} (25 trials). Similar ideas were recently developed for humanoid robots with QP-based controllers~\cite{spitz2017trial}.



\subsection{Simulation-to-reality \& meta-learning approaches}

The main idea behind meta-learning and \emph{SimToReal} approaches is to find a policy that is robust to a distribution of tasks (or environments). \emph{SimToReal} approaches exploit parameterized simulators in order to learn a policy that can effectively transfer on the real system. SimToReal algorithms can be categorized into ones that find policies that are robust: (1) to visual differences~\cite{sadeghi2017cad2rl,james2017transferring,james2018task,james2018sim} (\emph{domain randomization}), and (2) to different dynamics properties~\cite{chebotar2018closing,tan2018sim,peng2018sim} (\emph{dynamics randomization}).

James et al.~\cite{james2017transferring} use a rather simple controller, sample different goal targets and visual conditions (e.g., lighting, textures) and collect 1 million state-action trajectories of completing different goals. Once this dataset is collected, a convolutional NN, that will later serve as the policy, is trained in a supervised manner to find a mapping between image observations and the appropriate actions to take. Finally, they deploy this policy in the real world. Astonishingly, they were able to get $100\%$ success rate in the real world scenarios despite the fact that their task involved contacts and anticipating dynamic effects (i.e., picking and placing objects in a basket). 
Peng et al.~\cite{peng2018sim} use the Hindsight Experience Replay~\cite{andrychowicz2017hindsight} algorithm in order to maximize the expected return across a distribution of dynamics models. The dynamics parameters include masses and lengths of the links, damping and friction coefficients among others. Using their algorithm a 7-DOF manipulator learned how to push a puck on a desired location and directly transfered from simulation to reality.

However, these approaches do not provide any online adaptation capabilities; this basically means that if for some reason the policy does not generalize to the real world instance, the robot cannot improve its performance. 
SimOpt~\cite{chebotar2018closing} tries to close the loop by using real experience in order to find the distribution of the dynamics models to optimize on, but this type of approaches is very similar to model-based policy search with priors on the dynamics models (see Sec.~\ref{sec:priors_dynamics}). We can draw a parallel here and argue that model-based policy search with probabilistic models is performing something similar to dynamics randomization.
More concretely, performing policy search under an uncertain model is equivalent to finding a ``robust'' policy that can perform well under various dynamics models: the ones defined by the mean predictions and the uncertainty of the model.
%

Similarly, meta-learning approaches~\cite{feurer2015initializing, finn2017model, clavera2018learning, saemundsson2018meta} do not only try to find a robust policy but also a learning rule that can allow for fast adaptation (i.e., good performance with few gradient steps). Model-Agnostic Meta-Learning (MAML)~\cite{finn2017model} learns a good set of initial policy parameters, $\boldsymbol{\theta}_0$, such that every task can be solved within few gradient steps. 
A few applications of meta-learning target fast robot adaptation with promising results~\cite{clavera2018learning,saemundsson2018meta}. For example, S{\ae}mundsson et al.~\cite{saemundsson2018meta} model the distribution over systems using a latent embedding and model the dynamics using a global function (with GPs) conditioned on the latent embedding. They were able to learn control policies for the cartpole swing-up and the double pendulum tasks in less than $30\,s$ of interaction time including the meta-training time. Clavera et al.~\cite{clavera2018learning} use MAML to train a dynamics model prior such that, when combined with recent data, this prior can be rapidly adapted to the local context. They were able to combine their dynamics model with MPC in order to control a six-legged miniature physical robot in unknown/new situations (e.g., payload or different terrains), but still required $30$ minutes of interaction time for the meta-training process.









\begin{messagebox}{msg:expensive}
	Simulation-to-reality or meta-learning approaches can produce robust and adaptive policies that offer fast adaptation at test time. While they typically require expensive interaction time before the mission (e.g., in simulation), this should not be feared, as they can possibly produce the right prior for the task at hand. If they are combined with some on-line adaptation or model-learning~\cite{hwangbo2019learning}, they can learn effectively.

	\emph{Recommended readings:}~\cite{chebotar2018closing}~\cite{clavera2018learning}~\cite{saemundsson2018meta}
\end{messagebox}

\section{Challenges and Frontiers} \label{sec:challenges}

\subsection{Scalability} \label{sec:challenge_scalability}

Most of the works we described so far have been demonstrated with simple robots and simple tasks, such as the cartpole swing-up task (4D state space, 1D action space)~\cite{deisenroth2011pilco} or simple manipulators~(4D state space, 4D action space)~\cite{chatzilygeroudis2017black}. By contrast, humanoid robots have orders of magnitude larger state-action spaces; for example, the 53-DOF iCub robot~\cite{tsagarakis2007icub} has a state space of more than 100 dimensions (not counting tactile and visual sensors~\cite{maiolino2013flexible}). Most of the current micro-data approaches are unable to learn with such complex robots.

On the one hand, model-based policy search algorithms (Section~\ref{sec:model_based_ps}) generalize well to new tasks (since the model does not depend on the task) and learn high-dimensional policies with little interaction time (since the policy search happens within the model and not in interaction with the robot); but they do not scale well with the size of the state space: in the general case, the quantity of data to learn a good approximation of the forward model scales exponentially with the dimensionality of the state-space (this is the curse of dimensionality, see~\cite{bellman1957dynamic}). A factored state representation may provide the means to tackle such complexity, for example, by using dynamic Bayesian networks~\cite{dean1989model} to represent the forward model~\cite{boutilier2000stochastic}, but we are not aware of any recent work in this direction.

On the other hand, direct policy search algorithms (Sections~\ref{sec:imitation} and~\ref{sec:model_return}) can be effective in learning control policies for high-dimensional robots, because the complexity of the learning problem mostly depends on the number of parameters of the policy, and not on the dimensionality of the state-space; however, they do not generalize well to new tasks (when there is a model, it is specific to the reward) and they require a low-dimensional policy. Such a low-dimensional policy is an important, task-specific prior that constrains what can be learnt. For example, central pattern generators can be used for rhythmic tasks such as locomotion~\cite{ijspeert2008central}, but they are unlikely to work well for a manipulation task; similarly, quadratic programming-based controllers (and in general model-based controllers) can facilitate learning whole body controllers for humanoid robots~\cite{spitz2017trial,kumar2018improving}, but they impose the control strategy and the model.

In summary, model-based policy search algorithms scale well with the dimensionality of the policy, but they do not scale with the dimensionality of the state space; and direct policy search algorithms scale well with the dimensionality of the state-space, but not with the dimensionality of the policy. None of these two approaches will perform well on every task: future work should focus on either scaling model-based policy search algorithms so that they can learn in high-dimensional state spaces, or scaling direct policy search algorithms so that they can use higher-dimensional policies.


The dimensionality of the sensory observations is also an important challenge for micro-data learning: to our knowledge, no approach that performs ``end-to-end learning'', that is, learning with a raw data stream like a camera, has the efficiency of micro-data learning. Deep RL has recently made possible to learn policies from raw pixel input~\cite{mnih2015human}, largely because of the prior (i.e., an architectural inductive bias) provided by convolutional networks. However, deep RL algorithms typically require a very large interaction time with the environment (e.g., 38 days of play for Atari 2600 games~\cite{mnih2015human}), which is not compatible with most robotics experiments and applications. To address this challenge, a potential starting point is to use unsupervised learning to learn low-dimensional features, which can then be used as inputs for policies. Interestingly, it is possible to leverage priors to learn such state representations from raw observations in a reasonable interaction time~\cite{jonschkowski2015learning,lesort2018state}. It is also possible to create forward models in image space, that is, predicting the next image knowing the current one and the actions, which would allow to design model-based policy search algorithms that work with an image stream~\cite{oh2015action,ha2018world,assael2015data,eslami2018neural}.

\subsection{Priors} \label{sec:challenge_priors}

\begin{figure}
	\centering
	\includegraphics[width=0.5\linewidth]{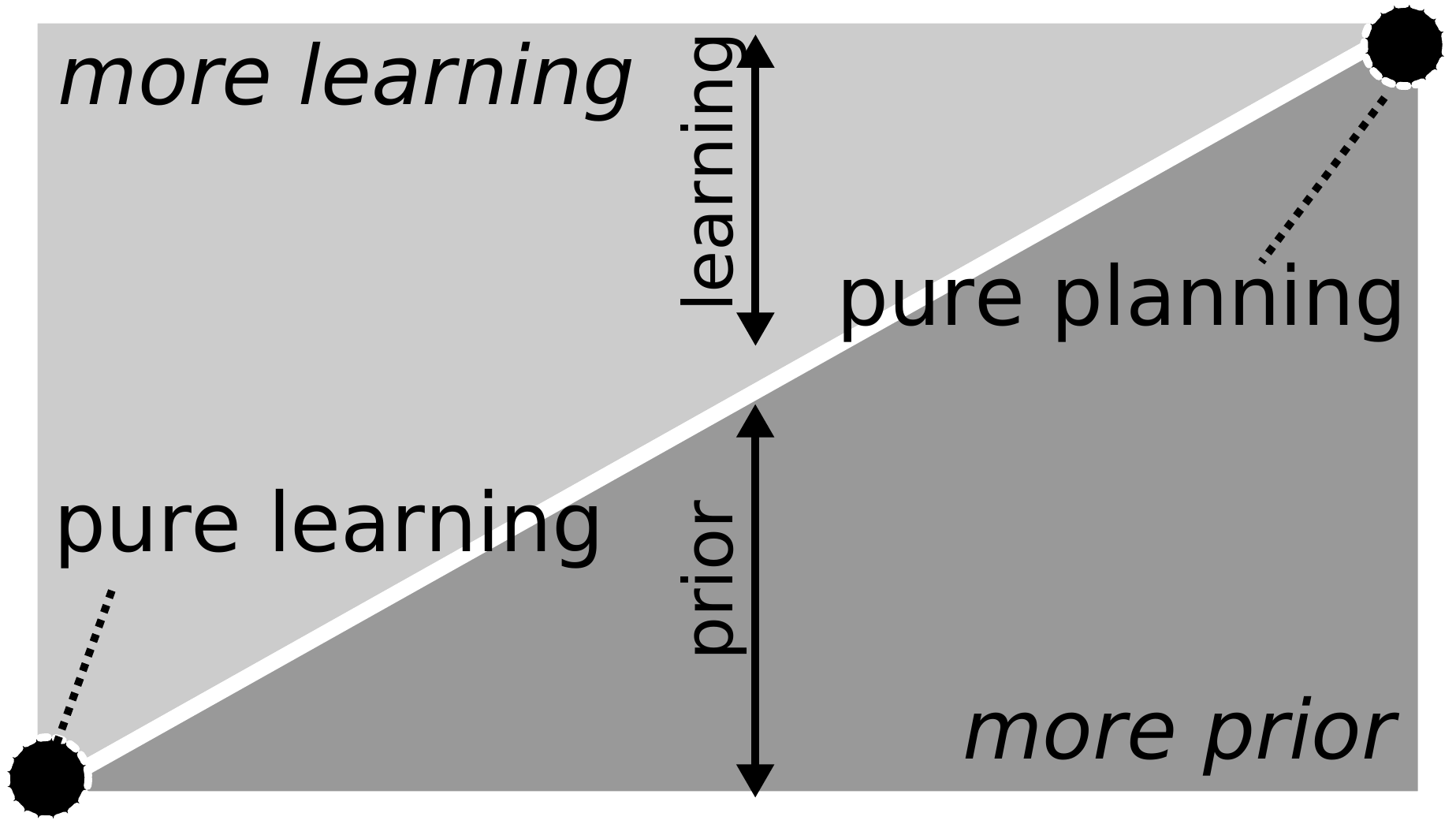}
	\caption{\label{fig:priors}The trade-off between prior knowledge and learning: for any task, there is an infinity of combinations between the amount of prior knowledge and the amount of learning required (image based on a slide by Oliver Brock, 2017).}
\end{figure}
Evolution has endowed animals and humans with substantial prior knowledge. For instance, hatchling turtles are prewired to run towards the sea~\cite{musick1997habitat}; or marine iguanas are able to run and jump within moments of their birth in order to avoid being eaten by snakes\footnote{As portrayed in the recent documentary ``Planet Earth 2'' from BBC.}. These species cannot rely on online learning mechanisms for mastering these behaviors: without such priors they would simply cease to exist.

Similarly to priors obtained from nature, artificial agents or robots can learn very quickly when provided with the right priors, as we presented in Sections~\ref{sec:priors_policy}, \ref{sec:priors_return}, and \ref{sec:priors_dynamics}. In other words, priors play a catalytic role in reducing the interaction time of policy search methods. Thus, the following questions naturally arise (Fig.~\ref{fig:priors}): what should be innate and what should be learned? and how should the innate part be designed?

Most of the existing methodologies use task-specific priors (e.g., demonstrations). Such priors can greatly accelerate policy search, but have the disadvantage of requiring an expert to provide them for all the different tasks the robots might face. More generic or task-agnostic priors (e.g., properties of the physical world) could relax these assumptions while still providing a learning speedup. Some steps have been made into identifying such task-agnostic priors for robotics, and using them for state representation~\cite{jonschkowski2015learning, lesort2017unsupervised}. We believe this is an important direction that requires more investigation. Meta-learning~\cite{feurer2015initializing, finn2017model, clavera2018learning, saemundsson2018meta} is a related line of work that can provide a principled and potentially automatic way of designing priors.

Physical simulations can also be used to automatically generate priors while being a very generic tool~\cite{cully_robots_2015,pautrat2018bayesian, antonova2017deep, antonova2016sample}. By essence, physical simulations can run in parallel and take advantage of faster computing hardware (from clusters of CPUs to GPUs): learning priors in simulation could be an analog of the billions of years of evolution that shaped the learning systems of all the current lifeforms.


While priors can bootstrap policy search, they can also be misleading when a new task is encountered. Thus, an important research avenue is to design policy search algorithms that can not only incorporate well-chosen priors, but also ignore those that are irrelevant for the current task~\cite{chatzilygeroudis2018using}. Following this line of thought, a promising idea is to design algorithms that actively select among a variety of priors~\cite{pautrat2018bayesian}.

\subsection{Generalization and robustness} \label{sec:challenge_generalization}

The majority of aforementioned articles are not much concerned with the generalization abilities and the robustness of the learned policy: they are designed to solve a single task, in a single context, with often little evaluation of the abilities to reject perturbations. For example, IT\&E~\cite{cully_robots_2015} focused on a repertoire of forward walking gaits for a hexapod robot on flat ground, rather than on various surfaces (e.g., incline surface or stairs) or various directions~\cite{chatzilygeroudis2018reset}; PILCO was applied for stacking a tower of foam blocks with a robotic manipulator~\cite{deisenroth_gaussian_2015}, but the task remained fixed over the course of learning (e.g., the size of the cubes did not vary) and there were no external perturbations (e.g., a wind gust). Put differently, in most of the reported experiments, the algorithms are very likely to have ``overfitted'' the robot and the task.

This situation could appear surprising because generalization and robustness are two of the most important questions in machine learning and control theory~\cite{siciliano2016springer}. Its source is, however, straightforward: assessing the robustness or the generalization abilities of a policy typically requires a significant additional interaction time. For example, a typical approach to measure the robustness of a control policy is to evaluate it with many different starting conditions and perturbations; a similar technique is often used to test the generalization abilities. Nevertheless, such an approach multiplies the interaction time by the number of tested conditions, which is likely to make the algorithm very quickly intractable on a real robot. In addition, this problem is amplified when the dimension of the state space increases, since there exist many more ways of perturbing a high-dimensional system than a low-dimensional one.

A potential remedy is to use policies that are intrinsically robust to some perturbations, that is, designing the policy space such that a change in the parameter space keeps the policy robust. For instance, the learning algorithm could search for a trajectory and a controller could be designed to follow it in a robust way: this corresponds to traditional trajectory optimization (or planning) in robotics~\cite{siciliano2016springer}. This is one of the ideas behind dynamic movement primitives (see Section~\ref{sec:priors_policy}), which act like ``attractors'' towards a trajectory of a fixed point. Similarly, it is possible to learn waypoints~\cite{lober2016efficient} or ``repulsors''~\cite{spitz2017trial} to mix learning with advanced, closed-loop ``whole-body'' controllers. It is, also, possible to incorporate optimization layers (e.g., a QP program~\cite{pham2018optlayer}) in a NN in order to take advantage of the structure they provide. Lastly, one can learn distinct \emph{soft} policies for simpler tasks and then compose them in order to achieve a more complicated task~\cite{haarnoja2018composable}.

It is also conceivable to learn models of the generalization abilities~\cite{pinville2011promote}, although it has, to our knowledge, never been tested with real robots. In that case, a model is trained to distinguish between behaviors (or trajectories) that are likely to overfit from those that are likely to be robust. This model can then be used in a policy search algorithm (e.g., in a constrained BO scheme).

Ultimately, we would like to have robots that can learn to execute various tasks quickly under varying conditions. This means that they need to be able to generalize from their previous experience without requiring much interaction time when the task changes. Having a policy that generalizes well offers the benefit of very fast execution, as opposed to algorithms that perform planning~\cite{chatzilygeroudis2018reset} or model identification~\cite{chatzilygeroudis2018using}. This challenge of micro-data multitask learning can be decomposed into two challenges. The first is about learning quickly to achieve different goals (i.e., only the reward function changes between tasks, for example, a robot that needs to throw a dart at different specified targets), while the second challenge is about adapting quickly to changes in the dynamics (i.e., the reward function does not change, for example, a robot that needs to cover as much distance forward as possible while walking on grass and transitioning on slippery ground).

Learning to achieve multiple goals has been tackled by a variety of methods, from using goal-conditioned policies, both in model-free (e.g., \cite{dasilva2012learning, kober2012reinforcement, fabisch2014active, schaul2015universal, karkus2016factored, ha2016evolutionary, abdolmaleki2017contextual, finn2017model, zhu2017target, andrychowicz2017hindsight, rauber2017hindsight, ghosh2017divide, mankowitz2018unicorn}) and model-based settings (e.g., \cite{deisenroth2014multi, kupcsik2017model}), to creating behavioral repertoires (e.g., \cite{cully_robots_2015, chatzilygeroudis2018reset, vassiliades2017cvt}).
Fast adaptation to changing dynamics could be addressed through BO (e.g., \cite{pautrat2018bayesian, supratik2018aloq}), meta-learning (e.g., \cite{vassiliades2013toward, wang2016learning, clavera2018learning, saemundsson2018meta, harrison2018control}), model identification (e.g., \cite{chatzilygeroudis2018using, yu2017preparing}), or generally policies that are robust to changes in the dynamics (e.g., \cite{rajeswaran2016epopt, peng2018sim}).
\subsection{Interplay between planning, model-predictive control and policy search} \label{sec:interplay}

The data-efficiency of policy search algorithms like PILCO or Black-DROPS rises from the fact that they learn and use dynamical models (Section~\ref{sec:model_based_ps}). However, if we assume that the dynamical model is known or can be learnt, there is a large literature on control methods that can be used. So, is policy search the right approach in such a case?

A fundamental controller from control theory is the linear-quadratic regulator (LQR)~\cite{kalman1960new}, which is optimal when the the dynamics are linear and the cost function is quadratic. Systems with nonlinear dynamics can be tackled with LQR by linearizing them around the current state and action, however, other approaches can be used such as differential dynamic programming~\cite{mayne1966second, jacobson1970differential} and its simpler variant, the iterative linear-quadratic Gaussian algorithm~\cite{todorov2005generalized} (iLQG). Generally, these methods can be used for optimal control with a large horizon lookahead, however, doing so can be computationally costly. For this reason, they are mostly employed to calculate trajectories offline; for example, GPS uses iLQG as the trajectory optimization procedure.

A way to permit online trajectory optimization is by reducing the horizon lookahead, thus, gaining in computational efficiency. This is known as model-predictive control (MPC)~\cite{garcia1989model}. Using shorter horizons, MPC is no longer optimal with respect to the overall, high-level task. This means that MPC can be used for short-term tasks, such as tracking a trajectory, which can be produced offline. The advantage of MPC is that it can get feedback from the real system and replan at every step. Such a control scheme can be very effective and has, for example, recently allowed real-time whole-body control of humanoid robots~\cite{koenemann2015whole}.

Although MPC can replan at every step, it still has the disadvantage of relying on models. Models can be inaccurate or wrong (especially in the first episodes of learning), therefore, there needs to be a mechanism that corrects the mismatch. A potential solution could be to combine iterative learning control~\cite{moore1992iterative, bristow2006survey} with MPC (e.g., see~\cite{lee1999model, lee2000model, wang2008iterative, assael2015data}). MPC additionally has the disadvantage of requiring full knowledge of the system state.
This problem can be mitigated by combining MPC with policy search. For example, in~\cite{zhang2016learning}, the authors used MPC with full state information during training, to learn NN policies that do not require full state information (only raw observations) when deployed, and even run faster than MPC online.


Should we then learn a big NN policy for complex high-level tasks, such as a humanoid robot helping with the house chores? Firstly, we need to consider that such complex tasks require long planning horizons. Secondly, as the task becomes more complex, so could potentially the policy space. Even if we do not consider memory requirements, learning such tasks from scratch would be intractable, even in simulation. One way of addressing such complexity is by decomposing the high-level task into a hierarchy of subtasks. Sampling-based planners~\cite{karaman2011sampling, browne2012survey} could operate at the high to mid levels of the hierarchy, whereas MPC could operate at the mid to low levels. Furthermore, policy search (or other algorithms for optimal control) can be used to discover primitives which themselves are used as components of a higher-level policy (e.g., see~\cite{duarte2017evolution}) or a planning algorithm (e.g., see~\cite{clever2017cocomopl, chatzilygeroudis2018reset}).
\subsection{Computation time} \label{sec:comptime}

Micro-data learning focuses on the desirable property of reducing the interaction time. However, most articles purposefully neglect computation time because they assume that it will be tackled automatically with faster hardware in the future. Although this is possible, it is worth investigating how different algorithms can potentially be sped up for near real-time execution with today's hardware.

For illustration, PILCO (see Section~\ref{sec:model_based_ps}) is a very successful and data-efficient algorithm, but can be very computationally expensive when the state-action or policy space dimensionality increases~\cite{chatzilygeroudis2017black,wilson_using_2014} (e.g., Wilson et al.~\cite{wilson_using_2014} report that PILCO required 3 weeks of computation time for 20 episodes on a 3-link planar arm task) and cannot take advantage of multi-core architectures. Black-DROPS and Black-DROPS with GP-MI (see Section~\ref{sec:priors_dynamics}) can greatly reduce the interaction time and take advantage of multi-core architectures, but they still require a considerable amount of computation time (e.g., Black-DROPS with GP-MI required 24 hours on a modern 16-core computer for 26 episodes of the pendubot task~\cite{chatzilygeroudis2018using}).
Both approaches use GP models which have a complexity that is quadratic to the number of points when queried; this is clearly inefficient when millions of such GP queries (e.g., Black-DROPS performs around 64M~\cite{chatzilygeroudis2017black}) are performed in each episode.

On the other hand, IT\&E~\cite{cully_robots_2015} and ``robust policies'' (e.g., see~\cite{peng2018sim},~\cite{yu2017preparing},~\cite{supratik2018aloq},~\cite{rajeswaran2016epopt}) can practically run in real-time because the prior is pre-computed offline. This ``recipe'' is shared by recent meta-learning methodologies, such as~\cite{finn2017model}, that aim to learn an expressive policy that can be optimized online using a single gradient update.

This does not mean that the offline precomputation time should not be optimized. Algorithms such as IT\&E or the work in~\cite{peng2018sim} use a form of directed exploration to create such a prior. If, for example, random search were used, it would probably need orders of magnitude more computation time to create a prior of the same quality.

\vspace{-0.5em}
\section{Conclusions} \label{sec:conclusion}
Thanks to recent advances in priors, policy representations, reward modeling, and dynamical models, it is now possible to learn policies on robots in a few minutes of interaction time. These micro-data learning algorithms considerably expand the usefulness of learning on robots: with these algorithms, we can envision robots that adapt ``in front of our eyes''. These algorithms, nonetheless, face critical challenges, most notably to scale-up simultaneously to high-dimensional state spaces and high-dimensional policy spaces.

As guidelines for future work in the field, we propose 5 precepts that summarize the ``generic rules'' that govern most of the work published so far about micro-data learning:
\begin{enumerate}
	\item Leveraging prior knowledge is key for micro-data learning: it should not be feared. However, the prior knowledge should be as explicit and as generic as possible.
	\item Use as much data as possible from each trial (e.g., trajectory data, not only reward value): when data is scarce, every bit matters.
	\item Take the time to choose what to test next (active learning): computers are likely to become faster in the future, but physics will not accelerate; it is therefore a sensible strategy to trade data resources for computational resources. It is still desirable, but less critical on the long term, to design algorithms that are fast enough to run on embedded systems.
	\item Every estimate (or model) should come with a measure of its uncertainty: when very little data is available, models will never have enough data to be ``right'' for the whole search space; algorithms must take this fact into account and reason with this uncertainty.
	\item If needed, use expensive algorithms before the mission: since we mostly care about online adaptation, we can have access to time and resources before the mission (access to computing clusters, GPUs, etc.)
\end{enumerate}

Finally, we would like to give a few recommendations for practical usage of micro-data algorithms:
\begin{itemize}
	\item\textbf{Low-DOF robots:} For robots with less than $~10$ DOFs, model-based policy search algorithms should be the choice of the researcher. Algorithms like Black-DROPS~\cite{chatzilygeroudis2017black} and PILCO~\cite{deisenroth2011pilco} will operate within reasonable computation time and will learn in very few trials.
	\item\textbf{High-DOF robots:} For robots with higher dimensional state/action space but with low dimensional policy spaces, Bayesian optimization approaches will provide the best trade-off between computation time and learning convergence. If a prior model or simulator is available, algorithms like IT\&E~\cite{cully_robots_2015} and MF-ES~\cite{marco17virtualvsreal} should be on the front line of learning in just a few trials.
	\item\textbf{Complex robots:} For robots with higher dimensional state/action space and high dimensional policy spaces, model-based policy search with priors on the dynamics will provide the most data-efficient results at the expense of increased computation cost. Algorithms like Black-DROPS with GP-MI~\cite{chatzilygeroudis2018using} and VGMI~\cite{zhu2018fast} effectively exploit parameterized simulators and should be able to learn in a handful of trials even for complex robots.
	\item\textbf{Raw observations:} When the observation (or state) space is very high dimensional (e.g., visual input), \emph{SimToReal} methods combined with online adaptation (e.g., SimOpt~\cite{chebotar2018closing}) should provide the best results.
\end{itemize}
In all cases, a good policy space and initialization of the policy parameters (e.g., from demonstrations~\cite{billard2008robot}) will accelerate learning.

\section{Acknowledgements}


This project received funding from: the European Research Council (ERC) under the European Union's Horizon 2020 research and innovation programme (GA no. 637972, project ``ResiBots'');
the Helmholtz Association through the project ``Reduced Complexity Models'';
the European Commission through the projects H2020 AnDy (GA no. 731540) and MEMMO (GA no. 780684); the CHIST-ERA project ``HEAP'';
the European Union's Horizon 2020 research and innovation programme under grant agreement No 739578 complemented by the Government of the Republic of Cyprus through the Directorate General for European Programmes, Coordination and Development.


\IfClass{IEEEtran}{
	\bibliographystyle{IEEEtran}
	\bibliography{IEEEabrv,survey,survey_policy_priors}
}{
	\vskip 0.2in
	\setlength{\bibsep}{2pt plus 0.3ex}
	\sffamily
	\small
	\bibliography{survey,survey_policy_priors}
}
\IfClass{IEEEtran}{
	\vspace{2em}
	\begin{IEEEbiography}
		[{\includegraphics[width=1in,height=1.25in,clip,keepaspectratio]{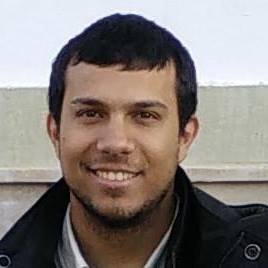}}]{Konstantinos Chatzilygeroudis}
		is currently a post-doctoral fellow at the LASA team at EPFL. He obtained a B.Sc. and M.Sc. in Computer Science and Engineering from the University of Patras in 2014, and a Ph.D. in Robotics and Machine Learning from Inria Nancy-Grand Est (France) and the University of Lorraine. His research interests lie in the area of artificial intelligence and focus on reinforcement learning and fast robot adaptation.~\\Website: \url{http://costashatz.github.io}
	\end{IEEEbiography}
	\vspace{2em}
	\begin{IEEEbiography}
		[{\includegraphics[width=1in,height=1.25in,clip,keepaspectratio]{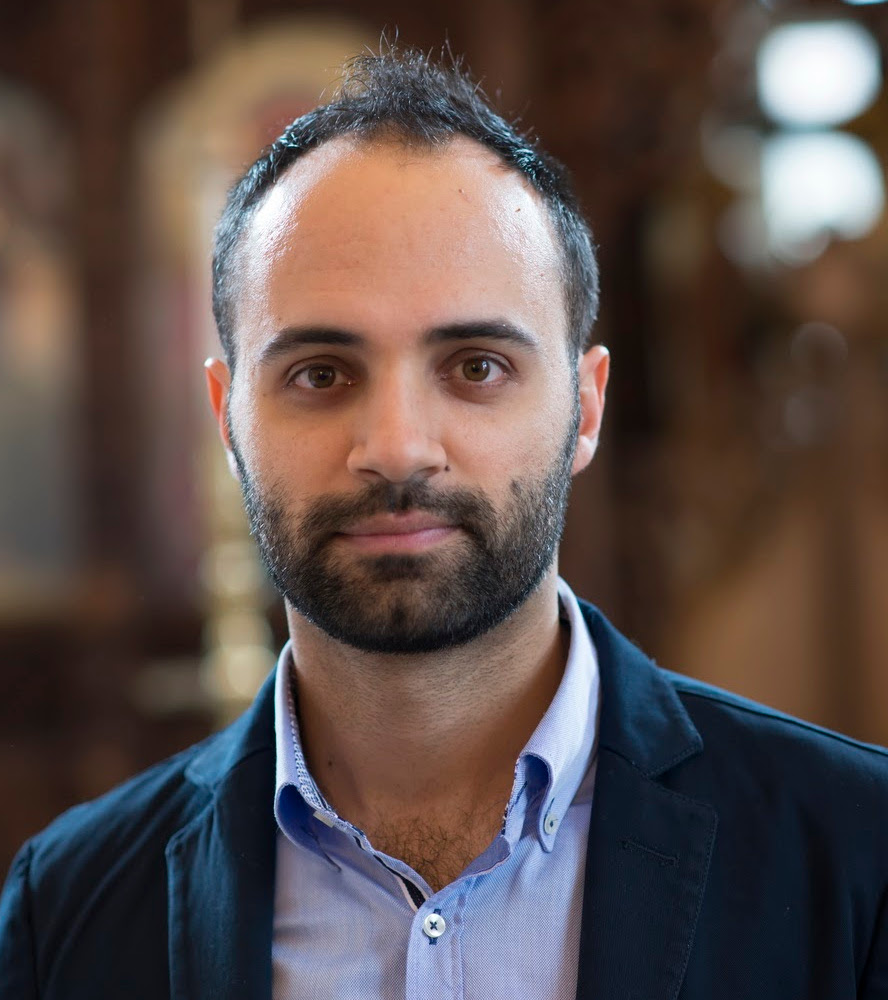}}]{Vassilis Vassiliades}
		received the Ph.D. degree from the University of Cyprus (2015). He is currently a team leader at the Research Centre on Interactive Media, Smart Systems and Emerging Technologies (RISE) in Cyprus. He held post-doctoral and research engineer positions at Inria Nancy, France (2015-2018), and research associate positions at the University of Cyprus (2015-2019) and RISE (2019). His research focuses on reinforcement learning, neural networks and evolutionary computation.
	\end{IEEEbiography}
	\vspace{2em}
	\begin{IEEEbiography}
		[{\includegraphics[width=1in,height=1.25in,clip,keepaspectratio]{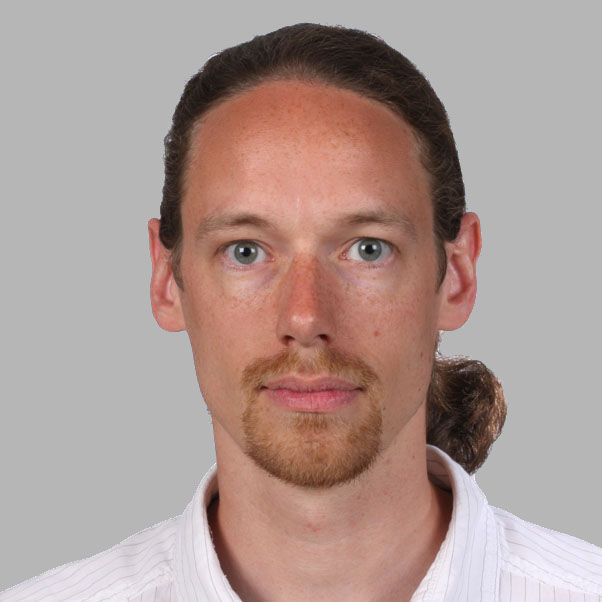}}]{Freek Stulp}
		received his doctorate degree in Computer Science from the Technische Universit\"{a}t M\"{u}nchen in 2007.
		He is currently the head of the department of Cognitive Robotics at the Institute of Robotics and Mechatronics at the German Aerospace Center (DLR).
		Previously, he was an assistant professor at the \'{E}cole Nationale Sup\'{e}rieure de Techniques Avanc\'{e}es (ENSTA-ParisTech).
		He currently serves as an Associate Editor in IEEE Transactions on Robotics.
	\end{IEEEbiography}
	\vspace{2em}
	\begin{IEEEbiography}
		[{\includegraphics[width=1in,height=1.25in,clip,keepaspectratio]{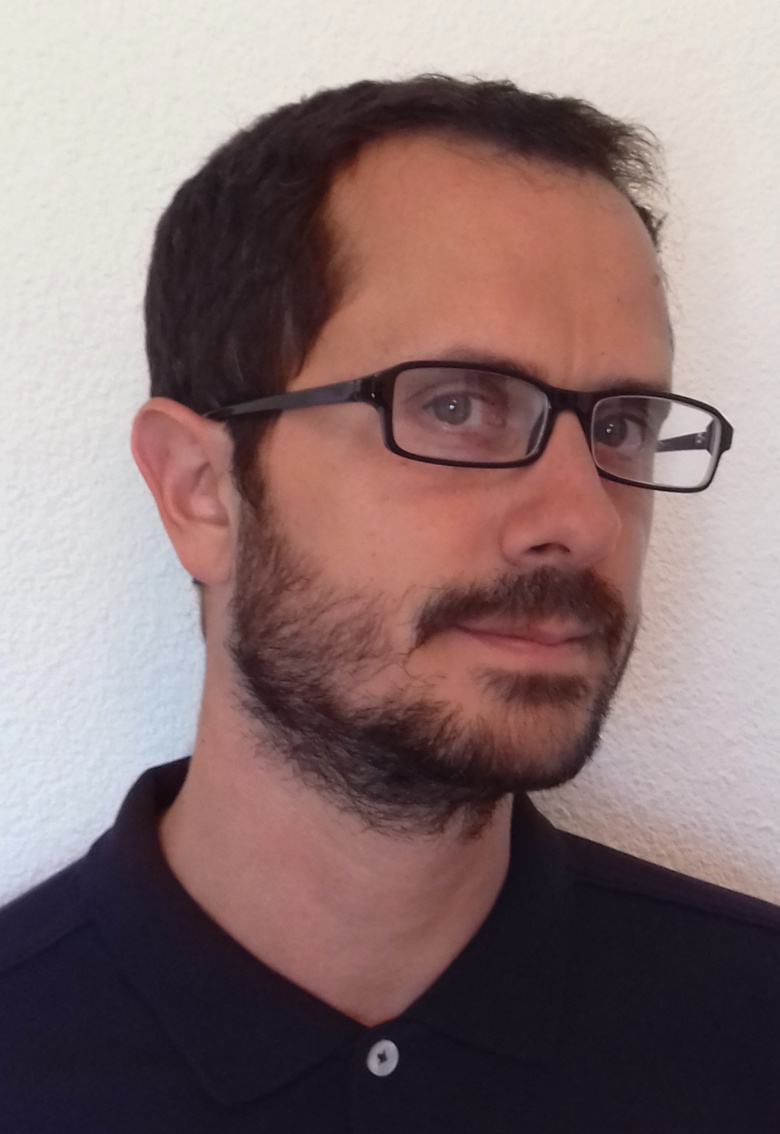}}]{Sylvain Calinon}
		received the Ph.D. degree from the Ecole Polytechnique F\'ed\'erale de Lausanne (EPFL) in 2007. He is a Senior Researcher at the Idiap Research Institute, and a Lecturer at the EPFL. From 2009 to 2014, he was a Team Leader at the Department of Advanced Robotics, Italian Institute of Technology. From 2007 to 2009, he was a Postdoc at EPFL. He currently serves as an Associate Editor in IEEE Transactions on Robotics and IEEE Robotics and Automation Letters. Website: \url{http://calinon.ch}
	\end{IEEEbiography}
	\vspace{2em}
	\begin{IEEEbiography}
		[{\includegraphics[width=1in,height=1.25in,clip,keepaspectratio]{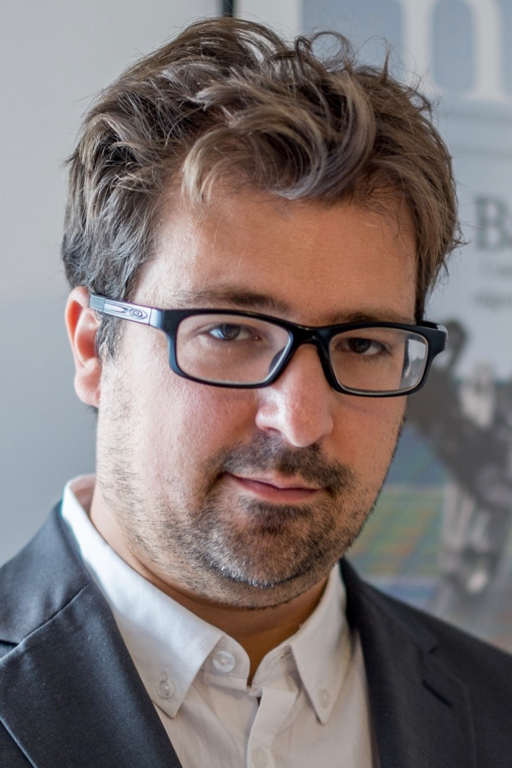}}]{Jean-Baptiste Mouret} received the Ph.D. degree in 2008 from the Pierre and Marie Curie University (Paris, France). He is currently a senior researcher (``Directeur de recherche'') at Inria, the French research institute dedicated to computer science and mathematics; from 2009 to 2015, he was an assistant professor (``ma\^itre de conf\'erences'') at the Pierre and Marie Curie University. His work was recently featured on the cover of Nature (Cully et al., 2015) and it received several national and international scientific awards, including the ``Prix La Recherche 2016'' and the ``Distinguished Young Investigator in Artificial Life 2017''. Website: \url{http://members.loria.fr/jbmouret}
	\end{IEEEbiography}
}{}

\end{document}